\documentclass{article}
\usepackage{graphicx}
\usepackage{multirow}
\usepackage{amsmath,amssymb} 

\usepackage{color}
\usepackage{url}
\usepackage{lscape}
\usepackage[numbers]{natbib}

\usepackage{algorithm}
\usepackage{algorithmic}
\usepackage[accepted]{icml2014}
\definecolor{lightgray}{gray}{.80}

\icmltitlerunning{Pain  Intensity Estimation by a Self--Taught Selection of  Histograms of
Topographical Features }

\begin{document}
\twocolumn[ \icmltitle{Pain  Intensity Estimation by a Self--Taught Selection of  Histograms of
Topographical Features  }

\icmlauthor{Corneliu Florea}{corneliu.florea@upb.ro} \icmladdress{Image Processing and Analysis Laboratory\\
                University "Politehnica" of Bucharest, Romania,
                Address Splaiul Independen\c{t}ei 313}

\icmlauthor{Laura Florea}{laura.florea@upb.ro} \icmladdress{Image Processing and Analysis Laboratory\\
                University "Politehnica" of Bucharest, Romania,
                Address Splaiul Independen\c{t}ei 313}

\icmlauthor{Raluca Boia}{rboia@imag.pub.ro} \icmladdress{Image Processing and Analysis Laboratory\\
                University "Politehnica" of Bucharest, Romania,
                Address Splaiul Independen\c{t}ei 313}
\icmlauthor{Alessandra Bandrabur}{abandrabur@imag.pub.ro} \icmladdress{Image Processing and Analysis Laboratory\\
                University "Politehnica" of Bucharest, Romania,
                Address Splaiul Independen\c{t}ei 313}
\icmlauthor{Constantin Vertan}{constantin.vertan@upb.ro} \icmladdress{Image Processing and Analysis Laboratory\\
                University "Politehnica" of Bucharest, Romania,
                Address Splaiul Independen\c{t}ei 313}

\icmlkeywords{Histograms of Topographical (HoT) features , Spectral Regression \sep Transfer
Learning, Temporal Filtering , Continuous Pain Intensity Estimation.
\end{keyword}}

\vskip 0.3in

]

\begin{abstract}
Pain assessment through observational pain scales is necessary for special categories of patients
such as neonates, patients with dementia, critically ill patients, etc. The recently introduced
Prkachin-Solomon score allows pain assessment directly from facial images opening the path for
multiple assistive applications. In this paper, we introduce the Histograms of Topographical (HoT)
features, which are a generalization of the topographical primal sketch, for the description of the
face parts contributing to the mentioned score. We propose a semi-supervised, clustering oriented
self--taught learning procedure developed on the emotion oriented Cohn-Kanade database. We use this
procedure to improve the discrimination between different pain intensity levels and the
generalization with respect to the monitored persons, while testing on the UNBC McMaster Shoulder
Pain database.
\end{abstract}


\section{Introduction}
In the past, the calculator was a mere tool for easing math. The rapid progress in the computer
science and in integrated micro-mechatronic, helped the appearance of assistive technologies. They
can improve the quality of life for all disabled, patients and elderly, but also for healthy
people. Assistive technologies include monitoring systems connected to an alarm system to help
caregivers while managing the activities associated with vulnerable people. Such an example is
automatic non-intrusive monitoring for pain assessment.

The International Association for the Study of Pain defines pain as ''an unpleasant sensory and
emotional experience associated with actual or potential tissue damage, or described in terms of
such damage'' \cite{PainBook:11}. Assessment of pain was showed to be a critical factor for
psychological comfort in the periods spent waiting at emergency units \cite{Gawande:10}. Typically,
the assessment is based primary on the self--report and several procedures are at hand; details can
be retrieved from \cite{Hugueta:10} and from the references therein. Complementary to the
self-report, there are observational scales for pain assessment and a review may be followed in
\cite{Baeyer:07}. If both methods are available, the self report should be the preferred choice
\cite{Shavit:08}.

Yet, there are several aspects that strongly motivate the necessity of the observational scales:
(1) Adult patients, typically, self-assess the pain intensity using a no-reference system, which
leads to inconsistent properties across scale, reactivity to suggestion, efforts at impressing unit
personnel etc. \cite{Hadjistavropoulos:04}; (2) Patients with difficulties in communication (e.g.
newborns, patients with dementia, patients critically ill) cannot self--report and assessment by
specialized personnel is demanded \cite{Baeyer:07}, \cite{Haslam:10}; (3) Pain assessment by nurses
encounters several difficulties. The third criteria is detailed by Manias et al. \cite{Manias:02}
by naming four practical barriers emerged from thorough field observations: (a) nurses  encounter
interruptions while responding to activities relating to pain; (b) nurses' attentiveness to the
patient cues of pain vary due to other activities related to the patients;  (c) nurses'
interpretations of pain vary with the incisional pain being the primary target of attention, and
(d) nurses' attempt to address competing demands of fellow nurses, doctors and patients. To respond
to these aspects, automatic appraisal of pain by observational scales is urged.

Among the multiple observational scales existing to the moment, the revised Adult Nonverbal Pain
Scale (ANPS-R) and the Critical Care Pain Observation Tool (CPOT) have been consistently found
reliable \cite{Stites:13}, \cite{Topolovec:13}, \cite{Chanques:14}. Both scales include evaluation
of multiple factors, out of which the first is the dynamic of the face expression. Intense pain is
marked by frequent grimace, tearing, frowning, wrinkled forehead (in ANPS-R) and, respectively,
frowning, brow lowering, orbit tightening, levator contraction and eyelid tightly closed (in CPOT).

The mentioned facial dynamics, in fact, overlap some of the action units (AU) as they have been
described  by the seminal Facial Action Coding Systems (FACS) introduced by \citeyear{Ekman:02}. A
practical formula to contribute to the overall pain intensity assessment from facial dynamics is
the Prkachin - Solomon formula \cite{Prkachin:08}. Here, the pain is quantized in 16 discrete
levels (0 to 15) obtained from the quantization of the 6 contributing face AUs :
\begin{equation}
    Pain = \begin{array}{l} AU_4 + \max\left(AU_6,AU_7\right) + \\
        \max\left(AU_9,AU_{10}\right) + AU_{43} \end{array}
    \label{Eq:PrkSolomon}
\end{equation}

The Prkachin - Solomon formula has the cogent merit of permitting direct appraisal of the pain
intensity from digital face image sequences acquired by regular video-cameras and image analysis.
Thus, it clears the path for multiple applications in the assistive computer vision domain. For
instance, in probably the most intuitive implementation \cite{Ashraf:09}, by means of digital
recording, a patient is continuously monitored and when an expression  of pain is detected, an
alert signal triggers the nurse's attention; he/she will further check the patient's state and will
consider measures for pain alleviation. Such a system may be employed in intensive care units,
where its main purpose would be to reduce the workload and increase the efficiency of the nursing
staff. Alternatively, it could be used for continuous monitoring of patients with communication
disabilities (e.g. neonates) and reduce the cost for permanent caring.

Following further developments (i.e. reaching high accuracy), in both computer vision and pain
assessment and management, automatic systems that use the information extracted from video
sequences could be applied to infer the pain intensity level and to automatically administer the
palliative care.

Another area of applicability is to monitor people performing physical exercises. For patients
recovering from orthopedic procedures, such an application would permit near real-time
identification of the movements causing pain, thus leading to more efficient adjustments of the
recovering program. For athletes or for normal persons training, such an application would
contribute to the identification of the weaker muscle groups and to fast improvement of the
training program.

In this paper we propose a system for face analysis and, more precisely, for pain intensity
estimation, as measured by the Prkachin--Solomon formula, from video sequences. We claim the
following contributions\footnote{This paper extends the work from \citeyear{Florea:14} by improving
the transfer method, by supplementary and more intensive testing and by adding filtering of the
temporal sequences and, thus boosting, the overall performance.}:
(1) we introduce the Histogram of Topographical (HoT) features that are able to address variability
in face images; (2) in order to surmount the limited number of persons, a trait typical for the
medical--oriented image databases, we propose a semi-supervised, clustering--oriented, self--taught
learning procedure; (3) we propose a machine learning based, temporal filtering to reduce the
influence of the blinks and to increase the overall accuracy; (4) we propose a system for face
dynamic analysis that applied to pain intensity estimation leads to qualitative results.


\subsection{Prior Art}
\label{Subsect:priorArt}

Although other means of investigation (e.g. bio-medical signals) were discussed \cite{Werner:13},
in the last period significant efforts have been made to identify reliable and valid facial
indicators of pain, in an effort to develop non-invasive systems. Mainly, these are correlated with
the appearance of three databases: the Classification of Pain Expressions (COPE) database
\cite{Brahnam:07} which focuses on infant classification of pain expressions, the Bio-Heat-Vid
\cite{Werner:13} database containing records of induced pain and the UNBC McMaster Pain Database
\cite{Lucey:11} with adult subjects suffering from shoulder pain. As said in the introduction, the
majority of the face--based pain estimation methods exploit the Action Unit (AU) face description,
previously used in emotion detection, and to which is correlated. A detailed review of the emotion
detection methods is in the work of Zeng et al. \cite{Zeng:09} and, more recently, in the work of
Cohn and De La Torre \cite{Cohn:14}.

On the COPE database, \citeyear{Brahnam:07} exploited Discrete Cosine Transform (DCT) for image
description followed by Sequential Forward Selection for reducing the dimensionality and nearest
neighbor classification for infant pain detection. On the same database, \citeyear{Gholami:10}
relied on relevance vector machine (RVM) applied directly on manually selected infant faces for
improved binary pain  detection. \citeyear{Guo:12} used Local Binary Pattern (LBP) and its
extension for improved face description and accuracy. We note that the COPE database, containing
204 images of 26 neonates is rather limited in extent and it is marked with only binary annotations
(i.e. pain and no-pain).

\citeyear{Werner:13} fused data acquired from multiple sources and information from a head pose
estimator to detect the triggering level and the maximum level of pain supportability, while
testing on the BioVid Heat Pain  database. One of their  contributions was to show that various
persons have highly different levels of pain triggers and of supportability levels, thus arguing
for pain assessment with multiple grades in order to accommodate personal pain profiles. At the
moment of writing this paper, the database is not public yet.

The pain recognition from facial expressions was referred in the work of \citeyear{Littlewort:07},
who applied a previously developed AU detector complemented by Gabor filters, AdaBoost and Support
Vector Machines (SVM) to separate fake versus genuine cases of pain; their work is based on AUs,
thus anticipating the more recent proposals built in conjunction with the UNBC McMaster Pain
Database.

Thus, due to its size and the fact that it was made public with expert annotation, the UNBC
McMaster Pain Database is currently the factum dataset for facial based pain estimation. In this
direction, \citeyear{Lucey:12} used Active Appearance Models (AAM) to track and align the faces on
manually labelled key-frames and further fed them to a SVM for frame-level classification. A frame
is labelled as ``with pain'' if any of the pain related AUs found earlier by \citeyear{Prkachin:08}
to be relevant is present (i.e. pain score higher than 0). \citeyear{Chen:13} transferred
information from other patients to the current patient, within the UNBC database, in order to
enhance the pain classification accuracy over Local Binary Pattern (LBP) features and AAM landmarks
provided by \citeyear{Lucey:12}.
\citeyear{Chu:13} introduced an approach based on Kernel Mean Matching named Selective Transfer
Machine (STM) and trained for person-specific AU detection, that is further tested on pain
detection. \citeyear{Zen:14} and \citeyear{Sangineto:14} trained a person specific classifiers
augmented with transductive parameter transfer for expression detection with applicability in pain.

We note that all these methods focus on binary detection (i.e. pain/no pain) thus experimenting
only with the first level of potential applications. Furthermore, pain (i.e. true case) appears if
at least one of the AU from eq. (\ref{Eq:PrkSolomon}) is present, case which happens in other
expressions too. For instance, AU 9 and 10 are also associated with disgust \cite{Lucey:10}.
Another corner case is related to the binary AU 43 which signals the blink; obviously not all
blinks are related to pain and the annotation of the UNBC database acknowledges this fact.

Multi-level pain intensity is estimated by the methods proposed in \cite{Kaltwang:12} and
\cite{Rudovic:13}. \citeyear{Kaltwang:12} jointly used LBP, Discrete Cosine Transform (DCT) and AAM
landmarks in order to estimate the pain intensity either via AU or directly at a sequence level
processing. \citeyear{Rudovic:13} introduced a  Conditional Random Field that is further
particularized for the person and, for the expression dynamics and timing so to obtain increased
accuracy.

Given the mentioned possible confusion between pain and other expressions and, respectively, the
explicit findings from \cite{Werner:13} regarding person dependent pain variability and the
implicit assumption from the pain scales, which use multiple degree for pain intensity, our work
focuses on pain intensity estimation. A byproduct will be pain detection.

\begin{figure*}
    \centering
         \includegraphics[width=0.65\textwidth]{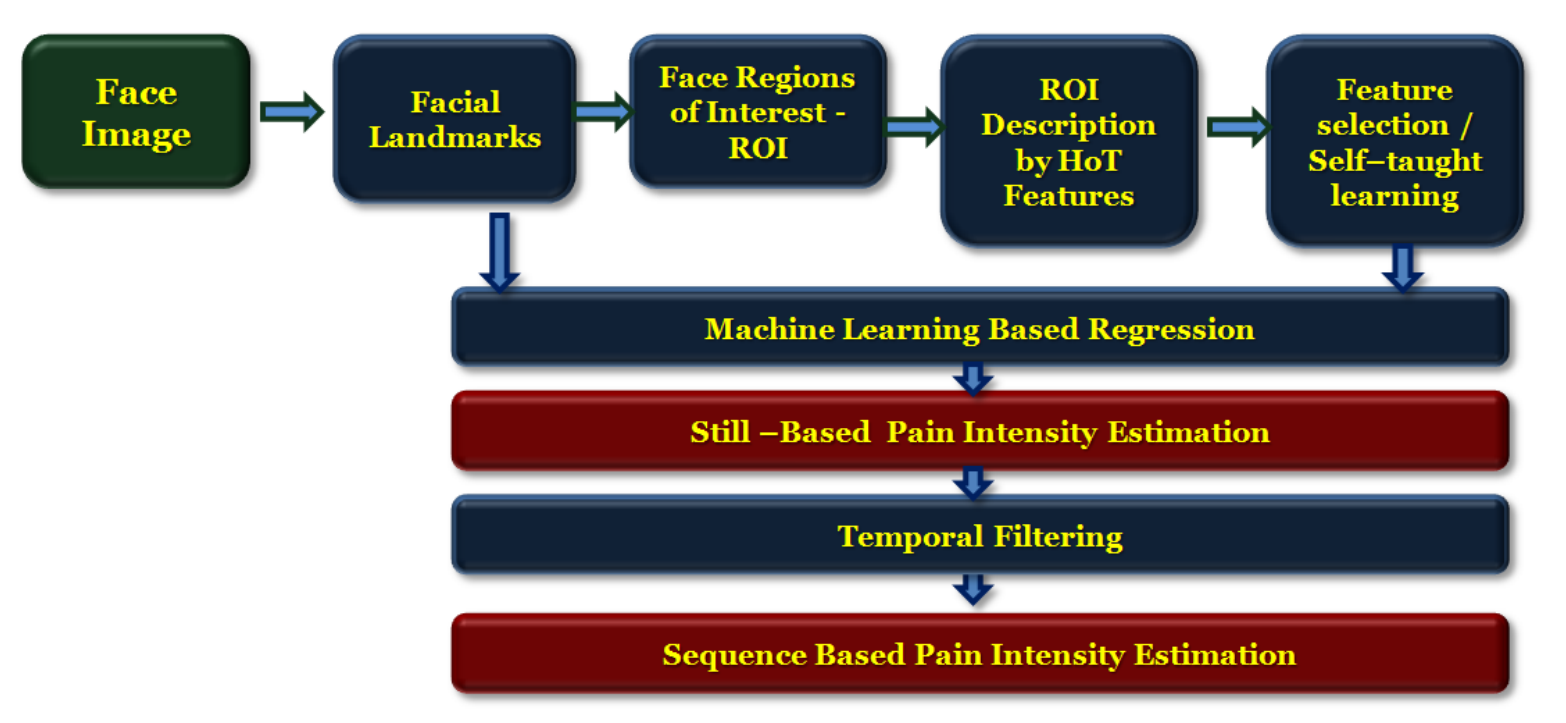}
    \caption{The schematic of the proposed continuous pain estimation method.
    \label{Fig:methodOverview} }
\end{figure*}

We propose a method working in a typical pattern recognition framework.  Given a face image and its
facial landmarks, out method will identify the regions of interest, that are further described by
Histogram of Topographical features. The important dimensions of the face description are selected
by a self-taught learning process that is followed by actual pain assessment via a machine learning
procedure. An overview of the proposed method is presented in figure \ref{Fig:methodOverview}.


\subsection{Paper Organization}
The remainder of the paper is structured as follows: in section \ref{Sect:Databases} we present the
used databases. In section \ref{Sect:Hot} we review state of the art feature descriptors and
introduce the here proposed Histogram of Topographical features. The procedure chosen for transfer
learning, as well as discussing alternatives, is presented in section \ref{Sect:Transfer}. The
system for still, independent, image--based pain estimation is presented in section
\ref{Sect:System}; we follow by the description of the temporal filtering of video sequences.
Implementation details and results are detailed in section \ref{Sect:Results}. The paper ends with
discussions and conclusions.


\section{Databases}
\label{Sect:Databases}

As mentioned, to our best knowledge, there exist three databases with pain annotations. The COPE
\cite{Brahnam:07} is rather small and with binary pain annotations, while the Bio-Heat-Vid
\cite{Werner:13} is to be made public. The  UNBC-McMaster Pain Database provides intensity pain
annotations for more than 48000 images.

\subsection{Pain Database}
\begin{figure*}
    \centering
         \includegraphics[width=0.75\textwidth]{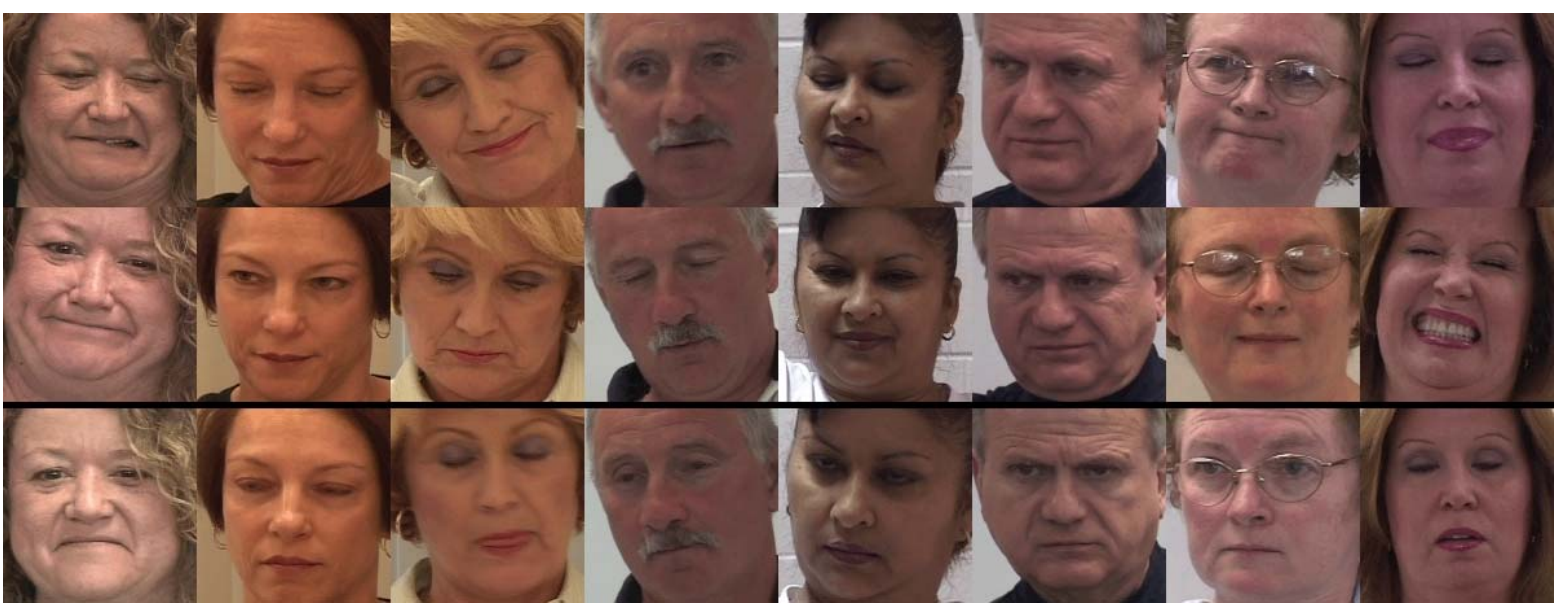}
    \caption{ Face crops from  UNBC-McMaster Shoulder Pain
    Expression Archive Database \cite{Lucey:11}. The top two rows illustrate the variability of pain
    faces while the bottom row illustrates non-pain cases. Note the similarity between the two
    situations.
    \label{Fig:UNBC_examples} }
\end{figure*}

We test the proposed system over the publicly available UNBC-McMaster Shoulder Pain Expression
Archive Database \cite{Lucey:11}. This database contains face videos of patients suffering from
shoulder pain as they perform motion tests of their arms. The movement is either voluntary, or the
subject's arm is moved by the physiotherapist. Only one of the arms is affected by pain, but
movements of the other arm are recorded as well, to form a control set. The database contains 200
sequences of  25 subjects, totalling 48,398 frames. One of the subjects lacks pain annotations and,
thus, it will be excluded from testing/training. Examples of pain faces proving the variability of
expressions is showed in figure \ref{Fig:UNBC_examples}.

The Prkachin - Solomon score for pain intensity is provided by the database creators, therefore
acting as a ground-truth for the estimation process. While in our work we do not focus on computing
separately the AUs, yet eq. (\ref{Eq:PrkSolomon}) explicitly confirms that databases build for AU
recognition are relevant for the pain intensity estimation.

The training testing scheme is the same as in the cases of \cite{Lucey:11} or \cite{Kaltwang:12}:
leave one person out cross validation; our choice is further motivated in section
\ref{Sect:Results}.

\subsection{Non Pain Database}
\begin{figure*}
    \centering
         \includegraphics[width=0.75\textwidth]{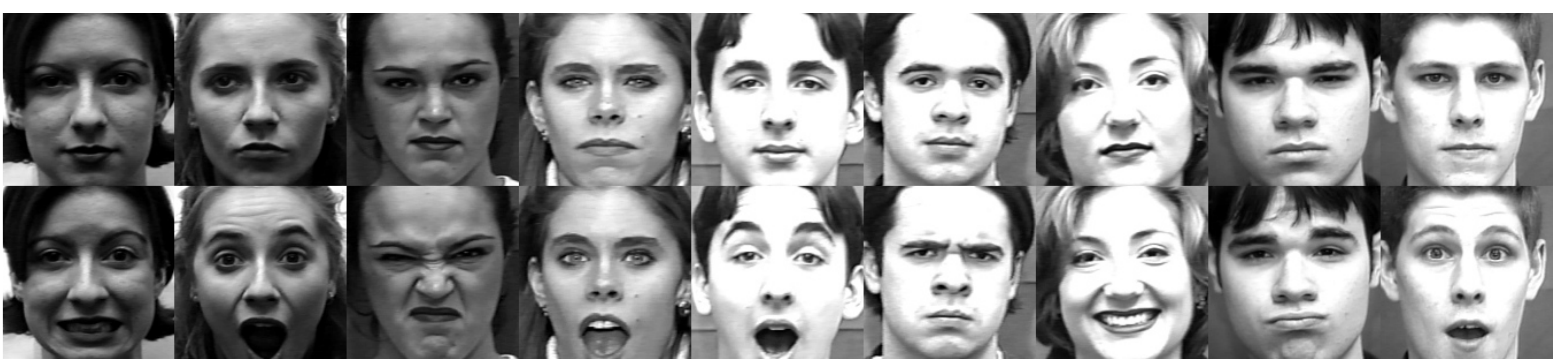}
          \caption{ Face crops pairs (neutral - top row  and respectively with expression bottom row) from
        the Cohn-Kanade database.
    \label{Fig:CK_Examples}}
\end{figure*}

Noting the limited number of persons available within the UNBC database (i.e. only 23 for the
training phase), we extend the data utilized for learning with additional examples from a non-pain
specific database, more precisely, the Cohn-Kanade database \cite{Kanade:00}. This contains 486
sequences from 97 persons and each sequence begins with a neutral expression and proceeds to a peak
expression. The peak expression for each sequence is coded in the FACS system thus having the AU
annotated. Relevant pairs of neutral/expression from the Cohn-Kanade database may be followed in
figure \ref{Fig:CK_Examples}.

\section{Histogram of Topographical Features}
\label{Sect:Hot}

To extract the facial deformation due to expression, we introduce a novel local/global descriptor,
namely the Histogram of Topographical (HoT) features. To proper place it in a context, we will
start by reviewing the most important image descriptors.

\subsection{Global/Local Image Descriptors - State of the Art.}

Many types of local image descriptors are used across the plethora of computer vision applications
\cite{Tuytelaars:08}. The majority of the solutions computed in the image support
domain\footnote{Here, alternatively to the image domain we assume the spectral domains where
popular descriptors such as DCT or wavelet coefficients are defined.} are approachable within the
framework of the Taylor series expansion of the image function, namely with respect to the order of
the derivative used.

Considering the zero-order coefficient of the Taylor series, i.e. the image values themselves, one
of the most popular descriptors is the histogram of image values and, respectively the data
directly, which was employed for instance in AAM \cite{Cootes:01} to complement the landmarks
shape. Next, relying on the first derivative (i.e. the directional gradient), several histogram
related descriptors such as HOG \cite{Dalal:05} or SIFT \cite{Lowe:04} gained popularity.

The second-order image derivative (i.e. the Hessian matrix) is stable with respect to image
intensity and scale and was part  of SIFT \cite{Lowe:04} and SURF \cite{Bay:08} image key-point
detectors. \citeyear{Deng:07} used the dominant eigenvalue of the Hessian matrix to describe the
regions in terms of principal curvature, while \citeyear{Frangi:98} deployed a hard classification
of the Hessian eigenvalues in each pixel (thus identifying the degree of local curviness) to
describe tubular structures (e.g. blood vessels) in medical images.

Summarizing, we stress that all the mentioned state of the art systems rely on information gathered
form a \emph{single} Taylor coefficient of either order zero, one or two in order to describe
images globally, or locally.

The approximation of the image in terms of the first two Taylor series coefficients is the
foundation of the topographical primal sketch introduced by \citeyear{Haralick:83} which is
inspired by the prior \citeyear{Marr:80} Laplacian based sketch representation. The primal sketch
was further adopted for face description by \citeyear{Wang:07}. In the primal sketch, the
description of the image is limited to a maximum number of 12 (or 16) classes which correspond to
the basic topographical elements. Further extension lays in the work of \citeyear{Lee:09}, who
plied the Hessian for locating key-points and described their vicinity with the histogram of color
values (order zero) and with the histogram of oriented gradients (order one).
\citeyear{Kokkinos:06} developed both the first and second derivative blob measures for an approach
derived from primal sketch features in terms of scale-invariant edge and ridge features; yet they
focus only on interest point and use different measures than our proposal.

In parallel to our work, \citeyear{Lindeberg:2014} proposes four strength measures extracted by
similarity with second order moment based Harris and Shi-Tomasi operator \cite{Shi:94}, but from
the Hessian's eigenvalues, that can be used to identify interest points.

We consider that all pixels from a region of interest carry important topographic information which
can be gathered in orientation histograms or in     normalized magnitude histograms. In certain
cases, only a combination of these may prove to be informative enough for a complete description of
images.


\subsection{Feature Computation}

In a seminal work, \citeyear{Haralick:83} introduced the so-called topographical primal sketch. The
gray-scale image is considered as a function $I:\mathbb{R}^2 \rightarrow \mathbb{R}$. Given such a
function, its approximation in any location $(i,j)$ is done using the second-order Taylor series
expansion:

\begin{equation}
    \label{Eq:taylorSeries}
    I(i+\Delta_i,j+\Delta_j) \approx
    \begin{array}{l}
     I(i,j) + \nabla I \cdot \left[ \Delta_i,\Delta_j \right] + \\
     \frac{1}{2} \left[%
    \begin{array}{cc} \Delta_i & \Delta_j \\ \end{array}\right] \mathcal{H}(i,j) \left[%
    \begin{array}{c} \Delta_i \\ \Delta_j \\ \end{array}\right] \end{array}
\end{equation}
where $\nabla I$ is the two-dimensional gradient and $\mathcal{H}(i,j)$  is the Hessian matrix.

Eq. (\ref{Eq:taylorSeries}) states that a surface is composed by a continuous component and some
local variation. A first order expansion uses only the $\nabla I$ term (the inclination amplitude)
to detail the  ''local variation'', while the second order expansion (i.e. the Hessian),
$\mathcal{H}(i,j)$, complements with information about the curvature of the local surface.
Considering the gradient and Hessian eigenvalues, a region could be classified into several primal
topographical features. This implies a hard classification and carries a limitation burden as it is
not able to distinguish, for instance, between a deep and a shallow pit. We further propose a
smoother and more adaptive feature set by considering the normalized local histograms extracted
from the magnitude of Hessian eigenvalues, the eigenvectors orientation and, respectively, the
magnitude and the orientation of the gradient.

\citeyear{Frangi:98} employed the concepts of linear scale space theory \cite{Iijima:62},
\cite{Florack:92}, \cite{Lindeberg:94} to elegantly compute the image derivatives. Here, the image
space is replaced by the scale space of an image $L(i,j,\sigma )$:

\begin{equation}
    L(i,j, \sigma) = G(i,j,\sigma) \ast I(i,j);
\end{equation}
where $\ast$ stands for convolutions and $G(i,j,\sigma)$ is a Gaussian rotationally symmetric
kernel with variance $\sigma^2$ (the scale parameter):

\begin{equation}
    G(i,j,\sigma) = \frac{1}{2\pi \sigma^2} e^{-(i^2+j^2)/2\sigma^2}
\end{equation}

The differentiation is computed by a convolution with the derivative of the Gaussian kernel:
\begin{equation}
\frac{\partial}{\partial i} L (i,j,\sigma) = \sigma I(i,j) \cdot \frac{\partial}{\partial i}
G(i,j,\sigma)
\end{equation}

In the scale space, the Hessian matrix $\mathcal{H}(i,j, \sigma)$ at location $(i,j)$ and scale
$\sigma$ is defined as:
\begin{equation}
    \mathcal{H}(i,j,\sigma) =
    \left(%
    \begin{array}{cc}
        L_{ii}(i,j,\sigma) & L_{ij}(i,j,\sigma) \\
        L_{ji}(i,j,\sigma) & L_{jj}(i,j,\sigma) \\
    \end{array}
    \right)
\end{equation}
where $L_{ii}(i,j,\sigma)$ is the convolution of the Gaussian second order derivative
$\frac{\partial^2}{\partial i^2}G(i,j,\sigma)$ with the image $I$ at location $(i,j)$, and
similarly for $L_{ij}(i,j, \sigma) = L_{ji}(i,j, \sigma)$ and $L_{jj}(i,j,\sigma )$. Further
analysis requires the computation of the eigenvalues and eigenvectors of the Hessian matrix.

The switch from the initial image space to the scale space, not only simplifies the calculus, but
the implicit smoothing reduces the noise influence over the topographic representation, influence
that was signaled as a weak point from inception by \citeyear{Haralick:83}.

\begin{figure*}[t]
    \begin{center}
    \begin{tabular}{cc}
        \includegraphics[width=0.18 \textwidth]{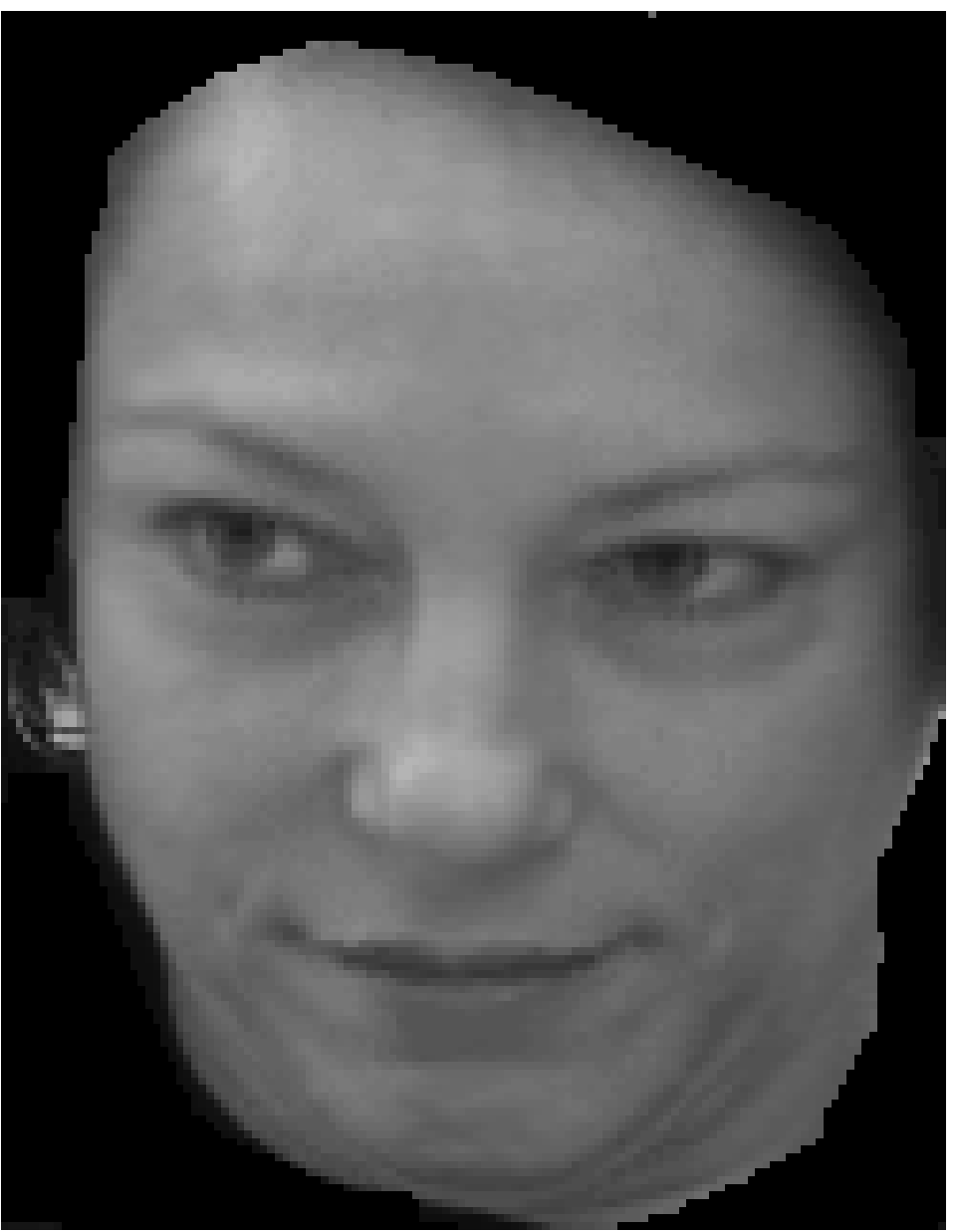} &
        \includegraphics[width=0.50  \textwidth]{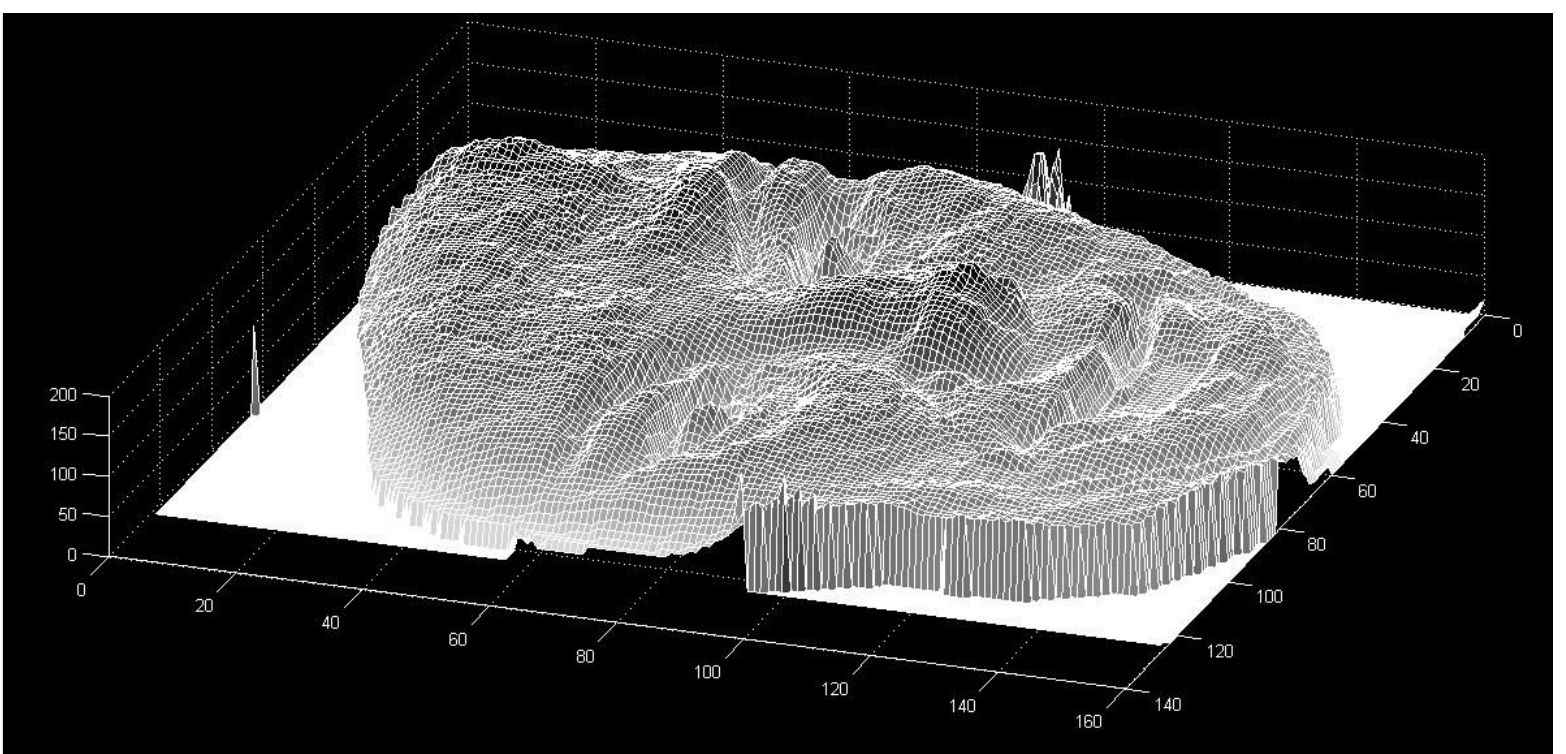} \\
        (a) & (b)
    \end{tabular}
    \begin{tabular}{cccc}
        \includegraphics[width=0.13 \textwidth]{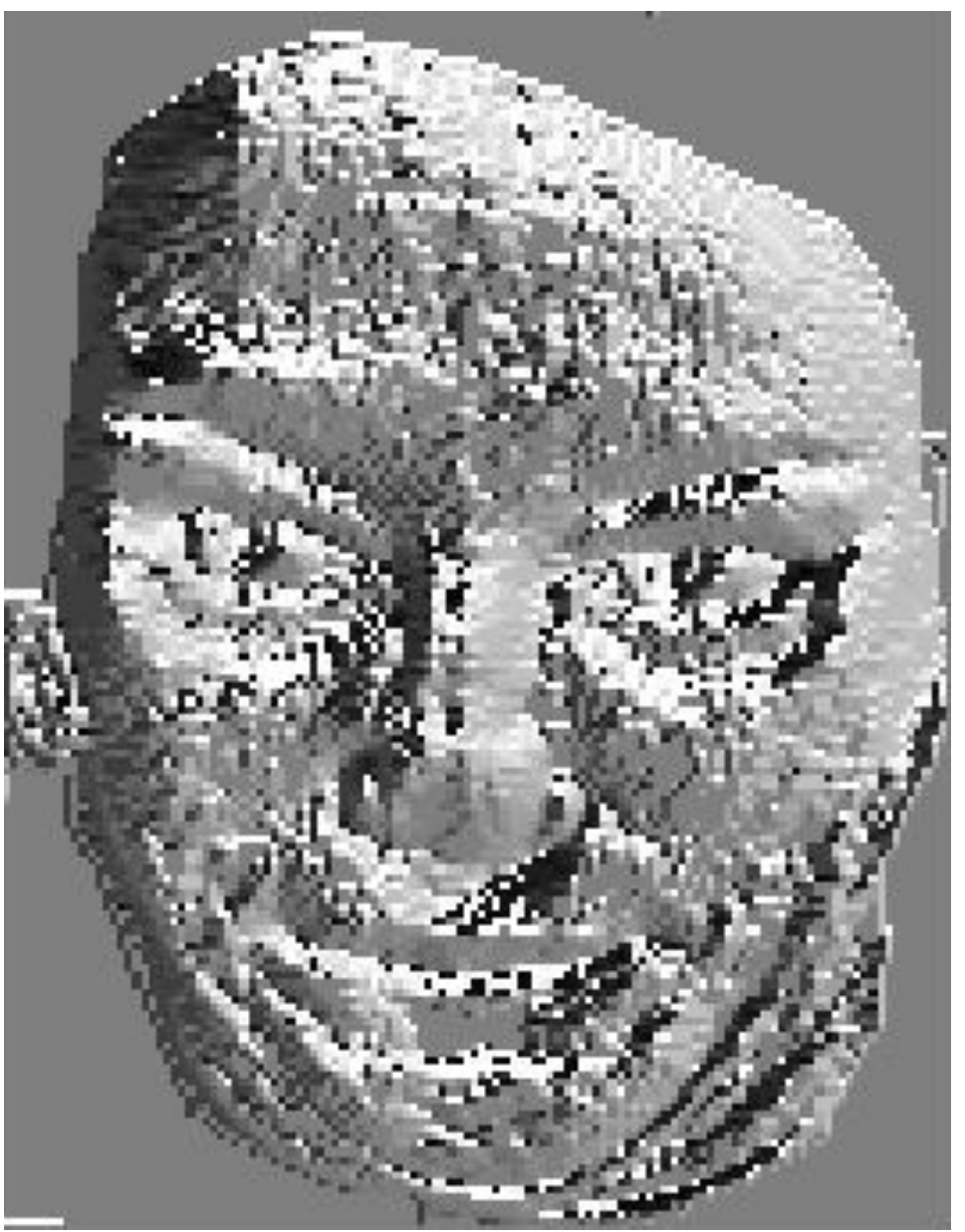} &
        \includegraphics[width=0.13 \textwidth]{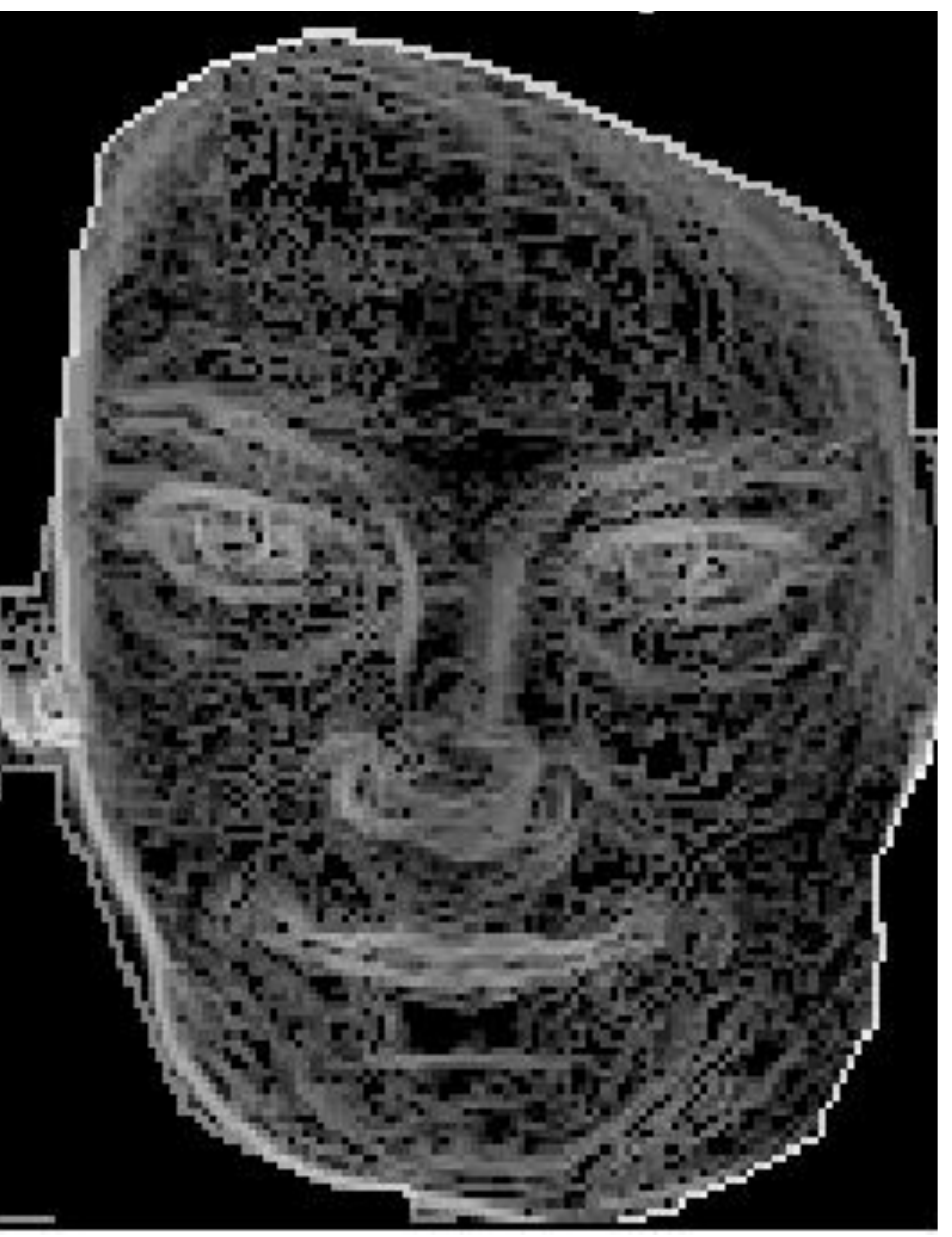} &
        \includegraphics[width=0.13 \textwidth]{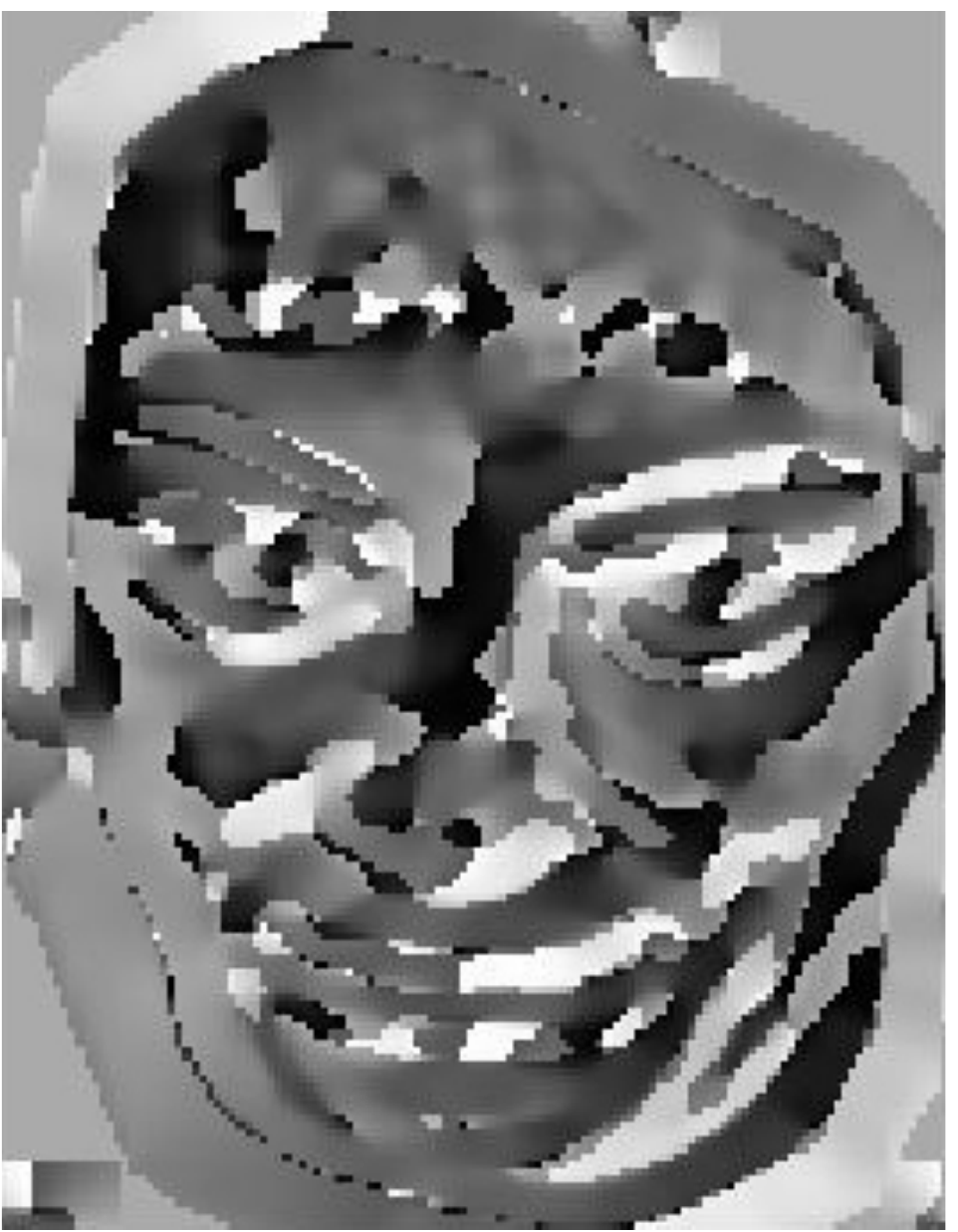} &
        \includegraphics[width=0.30 \textwidth]{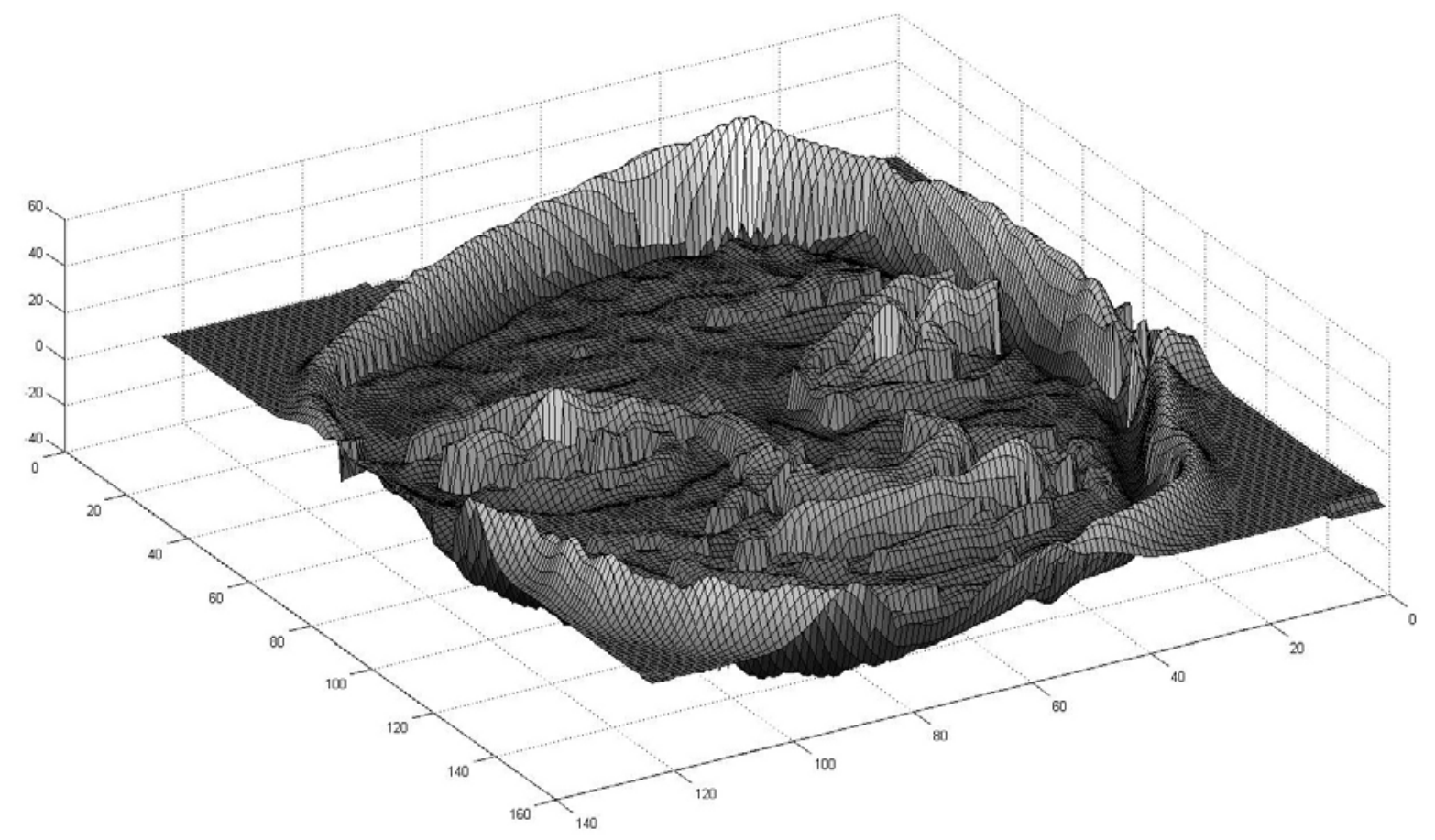} \\
        (c) & (d) & (e) & (f)
    \end{tabular}

    \end{center}
\caption{Computing the HoT features for a face: (a)  Original face image. (b)  The image
represented as a surface, (c) Gradient orientation image, (d) Gradient magnitude image, (e)
Curvature orientation image and (f) Curvature  strength image. \label{Fig:HoT_grad}}
\end{figure*}

The decomposition of the Hessian in eigenvalue representation acquiesce the principal directions in
which the local second order structure of the image can be decomposed. The second order hints to
the surface curvature and, thus, to the direction of the largest/smallest bending. We will denote
the two eigenvalues of the Hessian matrix $\mathcal{H}(i,j, \sigma)$ by $\lambda_1 (i,j,\sigma)
\leq \lambda_2 (i,j,\sigma)$. The eigenvector corresponding to the largest eigenvalue is oriented
in the direction of the largest local curvature; this direction of the principal curvature is
denoted by $\theta_{\lambda}(i,j,\sigma)$.  A visual example with gradient and curvature images  of
a face is shown in figure \ref{Fig:HoT_grad}.

\subsection{Descriptors for Regions of Interest }

In the remainder of the work, for each region of interest $\Omega$, the following HoT descriptors
will be used:
\begin{itemize}
    \item Second order data (Hessian):
    \begin{itemize}
        \item The histogram of hard voting of image surface curvature orientation. For each pixel in $\Omega$,
        ``1'' is added to the orientation of the ridge/valley extracted by computing the angle of the first
        Hessian eigenvector, if the second eigenvalue is larger than a threshold, $\lambda_2 > T_\lambda$.

        \begin{equation}
            \begin{array}{l}
            H_1^H \left( [ \theta ] \right) = \\
            \frac{1}{Z_1} \sum_{(i,j)\in \Omega} (\theta_{\lambda}(i,j)==[\theta])   \cdot
            (\lambda_2(i,j)>T_{\lambda})
            \end{array}
        \end{equation}

        \item The histogram of soft voting ridge orientation adds, instead of ``1'', the difference between the
        absolute values of the Hessian eigenvalues.

        \begin{equation}
          \begin{array}{l}
            H_2^H \left([\theta] \right) = \\
            \frac{1}{Z_2} \sum_{(i,j)\in \Omega} (\theta_{\lambda}(i,j)==[\theta]) \cdot
            (\lambda_2(i,j)-\lambda_1(i,j))
        \end{array}
        \end{equation}

        The $H_1^H$ and $H_2^H$ histograms produce, each, a vector of length equal to
        the number of orientation bins and describe the curvature strength in the image pixels.

        \item The range--histogram of the smallest eigenvalue, given a predefined range interval (e.g.
        $[0,M_{\lambda 2}=30]$).

        \begin{equation}
        \begin{array}{l}
            H_3^H ( k ) = \\
            \frac{1}{Z_3} \sum_{(i,j)\in \Omega} \left( \lambda_2(i,j)\in
            \left[ (k-1)\frac{M_{\lambda 2}}{N_{bin}}; \enspace k\frac{M_{\lambda 2}}{N_{bin}}
            \right] \right)
            \end{array}
        \end{equation}

        Inspired from the Shi-Tomasi operator \cite{Shi:94}, Lindeberg \cite{Lindeberg:2014} proposed
        to scan in that region the smaller Hessian eigenvalues and select the maximum of them as a measure of that
        region interest points. We differ by considering that not only the extremum of the minimum
        eigenvalues matters, but we gather all data in a histogram to have the region's global representation.

        \item The range--histogram of the differences between the eigenvalues given a predefined differences range interval (e.g.
        $[0,M_{\lambda 12}=50]$.

        \begin{equation}
          \begin{array}{c}
            H_4^H (k) = \frac{1}{Z_4} \sum_{(i,j)\in \Omega} \left( (\lambda_1(i,j) -
            \lambda_2(i,j))\in \right. \\
            \left. \left[ (k-1)\frac{M_{\lambda 12}}{N_{bin}};  \enspace  k\frac{M_{\lambda 12}}{N_{bin}}
            \right] \right)
            \end{array}
        \end{equation}

    \end{itemize}
    \item First order data (gradient):
    \begin{itemize}
    \item Histogram of orientation, $H_1^G$ \cite{Dalal:05}; each pixel having a gradient larger
    than a threshold, $T_G$ casts one vote;
    \item Histogram of gradient magnitude, $H_2^G$. The magnitudes are between 0 and a maximum value
    (100).
    \end{itemize}
\end{itemize}

The constants $Z_1,\dots, Z_4$ ensure that each histogram is normalized. Experimentally chosen
values for the thresholds are: $T_\lambda =0.1$ and $T_G = 5$. Each of the histograms is computed
on 8 bins.


\section{Modified Self--Taught Spectral Regression}
\label{Sect:Transfer}

The target database of the proposed system, UNBC, is highly extensive as number of frames, but is
also rather limited with respect to the  number of persons (only 25) and to inter-person
similarity. This is a typical trait of the medical--oriented image databases as there are not so
many ill persons to be recorded. To increase the robustness of the proposed algorithm, a new
mechanism for transfer learning is proposed.

We have inspired our work from the ``self--taught learning'' paradigm \cite{Raina:07} which is
conceptually similar to the inductive transfer learning \cite{Jialin:10}. A source database,
described by the unlabelled data $\mathbf{x}^{(1)}_u; \mathbf{x}^{(2)}_u; \dots ;$
$\mathbf{x}^{(k)}_u \in \mathbb{R}^n$ is used to learn the underlying data structure so to enhance
the classification over the labelled data of the target database: $\left\{ (\mathbf{x}^{(1)}_l;
y^{(1)}); (\mathbf{x}^{(2)}_l ; y^{(2)}); \dots ; (\mathbf{x}^{(m)}_l; y^{(m)}) \right\}$, where
$\mathbf{x}$ is the data and $y$ are labels. According to \cite{Raina:07}, the data structure could
be learned by solving the following optimization problem:

\begin{equation}
\begin{array}{r}
    minimize_{\mathbf{b},a}  \sum_i \left[ \| \mathbf{x}_u^{(i)} -  \sum_j a_j^{(i)}\mathbf{b}_j \|_2^2
    + \beta \|\mathbf{a}^{(i)}\|_1  \right]; \\
     s.t. \|\mathbf{b}_j\|_2  \leq 1, \forall j
     \end{array}
    \label{Eq:minRaina}
\end{equation}

The minimization problem from eq. (\ref{Eq:minRaina}) may be interpreted as a generalization of the
Principal Component Analysis concept\footnote{PCA is retrieved by solving $minimize_{\mathbf{b},a}
\sum_i \| \mathbf{x}_u^{(i)} - \sum_j a_j^{(i)}\mathbf{b}_j \|_2^2$ s.t. $\|\mathbf{b}_j\|_2=1$ and
$b_1,\dots b_T$ - orthogonal.} as it optimizes an overall representation, with the purpose of
identifying the best minimum set of linear projections. The PCA aims to decompose the original data
into a low-rank data and a small perturbation in contrast with Robust PCA \cite{Candes:11} which
decomposes the data into a low-rank sparse matrix.

Taking into account that the interest is in classification/regression, we consider that: 1. the
source database should be relevant to the classification task over the target database; 2. original
features should form relevant clusters such that, 3. the optimization over the source database
preserves local grouping. A modality to preserve the original data clustering is to compute the
Locality Preserving Indexing with the similarity matrix $\mathbf{W}$ using on the cosine distance:

\begin{equation}
    W_{i,j} = \left\{ \begin{array}{cl} \frac{\mathbf{x}_i^T \mathbf{x}_j}{\|\mathbf{x}_i \| \| \mathbf{x}_j \|
            } & \mathrm{if } \enspace \mathbf{x}_i \in N_p(\mathbf{x}_j) \vee \mathbf{x}_j \in N_p(\mathbf{x}_i) \\
            0 & otherwise
               \end{array}     \right.
\label{Eq:Similarit}
\end{equation}

We replaced the cosine distance used in \cite{Florea:14} with the heat kernel, as in the case of
Locality Preserving Projection \cite{He:03} with a further adaptation to our problem:

\begin{equation}
    W_{i,j} = \left\{ \begin{array}{cl} k e^{-\frac{ \| \mathbf{x}_i^T \mathbf{x}_j \|^2}{2\sigma^2}} &
                    \mathrm{if } \enspace \mathbf{x}_i \in N_p(\mathbf{x}_j) \vee \mathbf{x}_j \in N_p(\mathbf{x}_i) \\
            0 & otherwise
               \end{array}     \right.
\label{Eq:SimilaritHeat}
\end{equation}

where $N_p(\mathbf{x}_i)$ contains the $p=8$ closest neighbors of $\mathbf{x}_i$ and $k=1$ if $x_i$
contains at least one of the action units from eq. (\ref{Eq:PrkSolomon}). The optimization runs over
the similarity matrix, such that we solved the following regularized least squares problem over the
unlabelled source database:

\begin{equation}
\begin{array}{r}
    minimize_{\mathbf{B}=[\mathbf{b}_1 \dots \mathbf{b}_T]} \sum_i \left( \left( \mathbf{b}_j^T
     \mathbf{x}^{(i)}_u - u_i^j\right)^2 +  \alpha \|\mathbf{b_j}\|_2^2   \right);
     \\ i=1,\dots, k
     \end{array}
\label{Eq:SpectralRegress}
\end{equation}
where $u_i^j $ is the $j$-th element of the eigenvector $\mathbf{u}_i$ of the symmetrical
similarity matrix $\mathbf{W}$. This process of extracting the data representation (eq.
(\ref{Eq:SimilaritHeat}) - if removed the adaptation to our problem  and
(\ref{Eq:SpectralRegress})) form the so called \emph{spectral regression} introduced by
\citeyear{Cai:07}. A similar transfer learning method was proposed by \citeyear{Jiang:12}, with two
core differences: data similarity is computed using a hard assignment compared to the soft approach
from eq. (\ref{Eq:SimilaritHeat}) and unsupervised clustering was performed on the target database.

Finally, the labelled new data is obtained by classification of the projected vectors
$\mathbf{z}_l^{(i)}$, determined as:
\begin{equation}
    \mathbf{z}_l^{(i)} = \mathbf{B}      \mathbf{x}_l^{(i)}, \forall i=1,\dots m
    \label{Eq:Proj}
\end{equation}
where $\mathbf{B} =[\mathbf{b}_1 \dots \mathbf{b}_T]$.

In our algorithm, the neutral image and respectively the images with the apex emotion from
Cohn-Kanade database were the unlabelled data from the source database, while the UNBC was the
target, labelled, database. The transfer learning process and the projection equation,
(\ref{Eq:Proj}), were applied independently on the Hessian based histograms, $[H_1^H, \dots H_4^H]$
and, respectively, on the gradient based histograms $[H_1^G, H_2^G]$.

The transfer learning includes also a dimensionality reduction (i.e. feature selection procedure).
The full HoT feature has 240 dimensions, while there are 7937 images with Prkachin-Solomon score
higher than 0. Taking into account that part are utilized for training, feature selection is
required to prevent the classifier from falling into the curse of dimensionality. The Hessian based
histogram are reduced to $T_H=32$ dimensions while gradient ones to $T_G=24$.


\section{System}
\label{Sect:System}

\subsection{Still image pain estimation}
The schematic of the proposed system for pain intensity assessment in independent, still images is
presented in figure \ref{Fig:TransfKnowledge} (b). The procedure for HoT features extraction is
presented in figure \ref{Fig:TransfKnowledge} (a).

\subsection{Landmark localization and annotations}

The UNBC landmarks are accurate \cite{Lucey:11}, yet their information is insufficient to provide
robust pain estimation. In this sense, \citeyear{Kaltwang:12} reported that using only points, for
direct pain intensity estimation, a mean square error of 2.592 and a correlation coefficient of
0.363 is achieved (as also shown in table \ref{Tab:HoTResults}).

\begin{figure*}[t]
    \begin{center}
    \begin{tabular}{cc}
        \includegraphics[width=0.30 \textwidth]{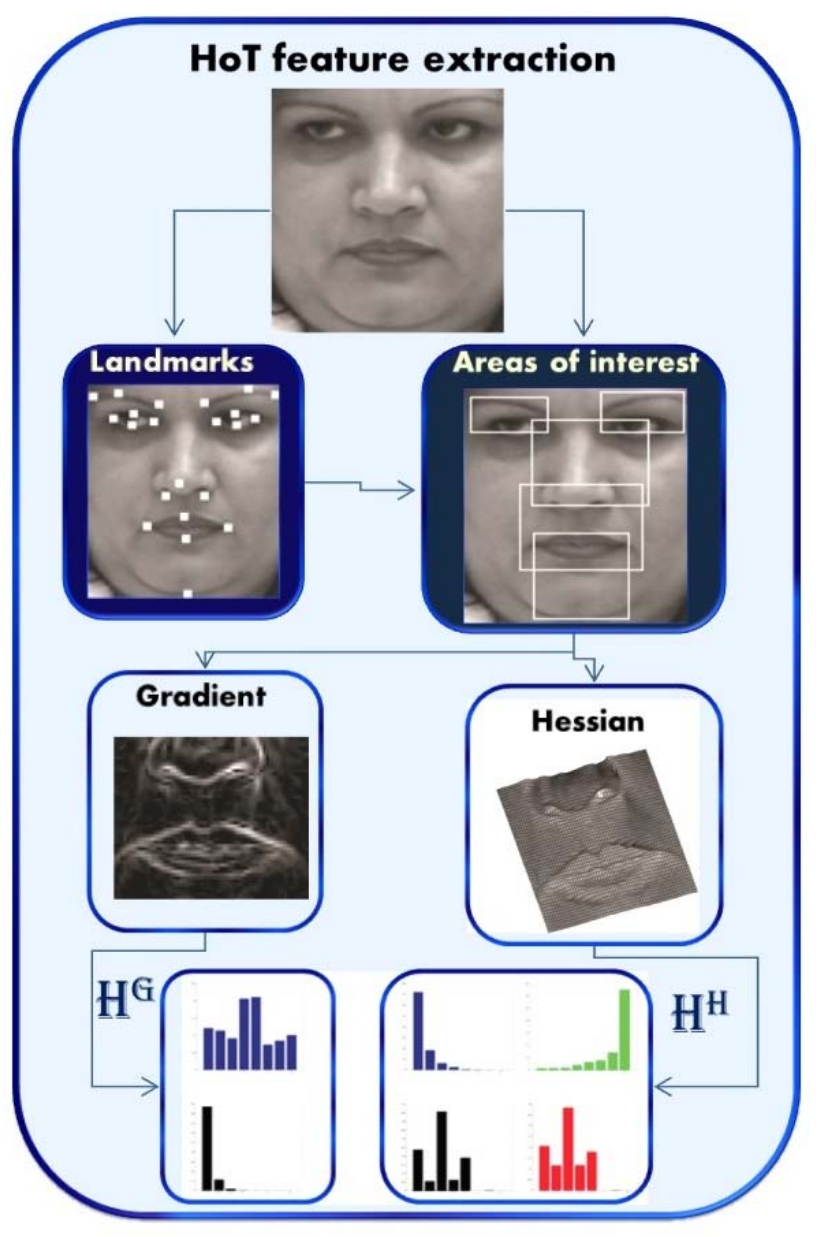} &
        \includegraphics[width=0.60 \textwidth]{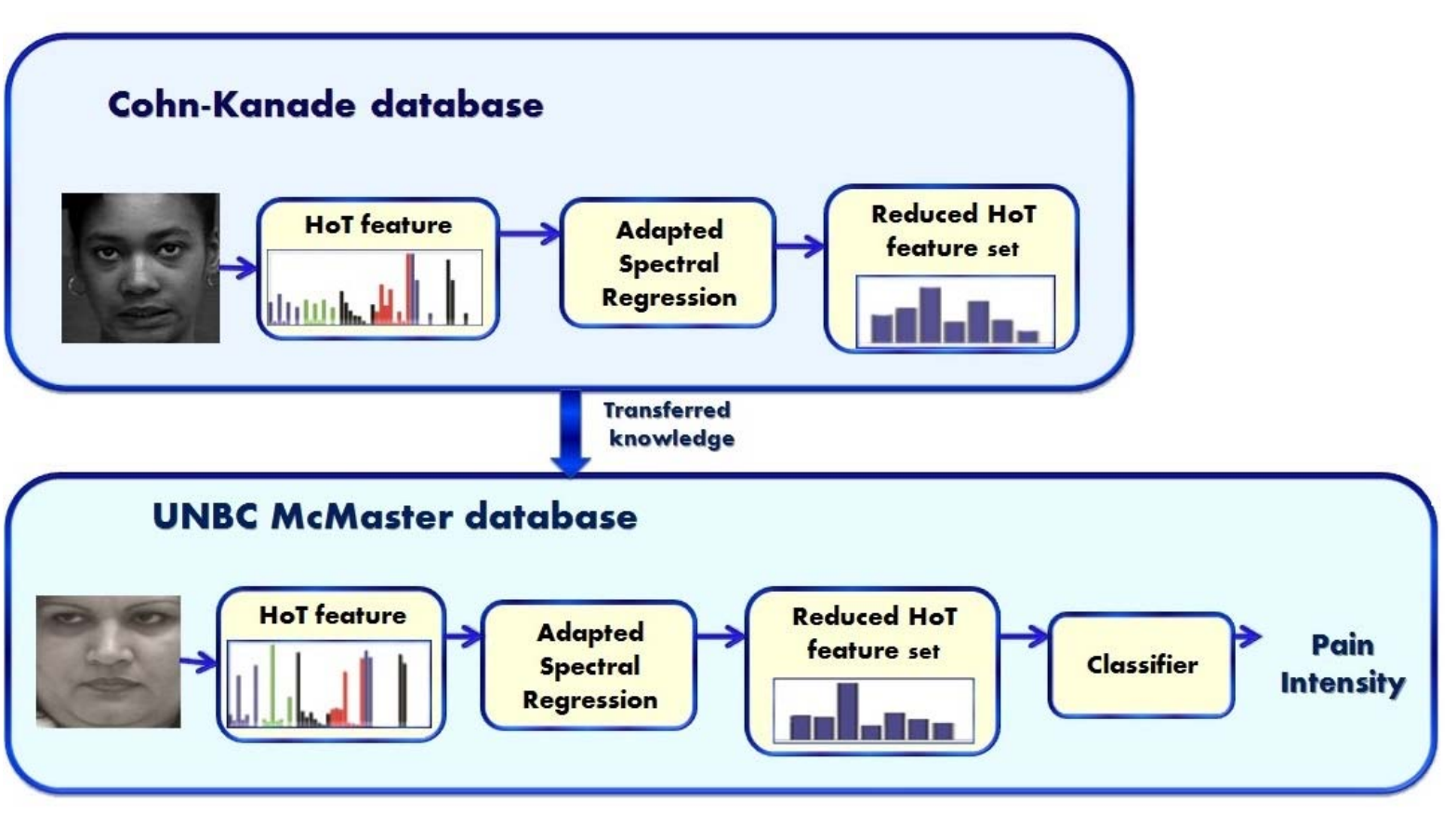} \\
        (a) & (b)
    \end{tabular}
    \end{center}
\caption{ (a)  The features extraction procedure. (b)  The transfer knowledge system. Data internal
representation is computed on unlabelled data from  Cohn-Kanade database to make use of the larger
number of persons. The reduced data is fitted in order to predict pain intensity.
\label{Fig:TransfKnowledge}}
\end{figure*}

Due to the specific nature of the AUs contributing to pain, and  based on the 22 landmarks, we have
selected 5 areas of interest, showed in figure \ref{Fig:TransfKnowledge} (a), as carrying
potentially useful data for pain intensity estimation.

Due to the variability of the encountered head poses, we started by roughly normalizing the images:
we ensured that the eyes were horizontal and the inter--ocular distance was always the same (i.e.
50). Out--of--plane rotation was not dealt with explicitly, but implicitly by the use of the
histograms as features. Since the 8 histogram bins span 360 degrees, the head robustness is up to
$22.5^0$.

\subsection{Temporal Filtering}
In our previous work on the topic \cite{Florea:14} and also in \cite{Kaltwang:12} it was acquainted
that while marked by equation (\ref{Eq:PrkSolomon}), the blink does not always signal pain.
Unfortunately, the blink is sufficiently obvious such that an automatic system concludes that it is
pain. The main difference between the blink and the pain face is duration: blinks typically take
less than 15 frames, while pain faces are longer. To further differentiate between those two, we
consider three versions of temporal filtering of the sequences.

The first solution is a simple filtering aimed at reducing the noise. Here we started with a median
filter on a vicinity of width $w$ followed by a linear regression (LR), over the same window to
estimate the current value. The preferred window size is $w=21$.

The second and third solutions rely on machine learning approaches, where given the data from the
vicinity of the current pixel, a classifier attempts to estimate a better value.

The difference between the two considered solutions lies in data description:
\begin{itemize}
    \item The feature vector is formed by the pain estimates for the frames in the vicinity $w$ taken from the sequence. One expects
    that given a large enough window size $w$, the classifier will learn to skip the blink. Here
    the only classifier that produces good results was a MLP with two hidden layers (of 40 neurons
    each) and single output. The MLP may be clearly seen as a generalization over the linear
    regression, in the sense that different feature dimensions contribute with different weights.

    \item The feature is obtained by considering statistical moments computed on increasing vicinities
     of still image pain estimates. With such description, we estimate that typically patterns that
     erroneously appear in the estimated data are learned and skipped in the testing.  Here, the feature of the frame $i$ is :
     \begin{equation}
        F (i) = \left[z_i, \sigma_{z_i^3}, \sigma_{z_i^5}, \dots  \sigma_{z_i^w},\right]
     \end{equation}
    where $z_i$ is the pain estimate for the frame $i$, while $\sigma_{z_i^w}$ is the
    variance of the pain estimates over a centered window in $i$ having the width $w$.
    This description is inspired from the total strict pixel
    ordering \cite{Coltuc:99}, \cite{Florea:07}. Again, we empirically found the best value for window
    size to be $w=61$.

     In this case a SVR leads to better correlation, while the 2-hidden layers MLP shows smaller mean
     square error.

\end{itemize}

The main idea behind these solutions is to gather data from vicinities larger than blink duration
and  to allow the classifier to distinguish between blinks relevant to pain and those which are not
relevant. Further more, we determined that still pain estimation produces patterns of estimates in
pain onset and offset and we aim to improve the performance in such cases.


\section{Results and Experiments}
\label{Sect:Results}

\subsection{Objective Metrics}
To objectively evaluate the performance of the proposed approach for the task of continuous pain
intensity estimation according to the Prkachin-Solomon formula, several metrics are at hand. The
mean squared error ($\overline{\varepsilon^2}$) and  the Pearson correlation coefficient ($\rho$)
between the predicted pain intensity and ground truth pain intensity are used for continuous pain
intensity accuracy appraisal. These measures were also used by \cite{Kaltwang:12}, thus direct
comparison is straight-forward.

For the pain detection all frames with Prkachin-Solomon higher than zero are considered with pain
and the measure adopted is Area Under ROC curved (AUC). While we argue against the relevance of
this method for the pain estimation in assistive computer vision, yet the measure is relevant to
evaluate the theoretical performance of a face analysis method.  The AUC was used also by several
other works \cite{Lucey:12}, \cite{Chen:13}, \cite{Chu:13}, \cite{Zen:14}, \cite{Sangineto:14}, and
it facilitates direct comparison with state of the art solutions.

\subsection{Testing and Training}

The used training-testing scheme, for both still and sequence related pain estimation is the
leave--one--person--out cross-validation. The same scenario is employed in previous works on the
topic \cite{Lucey:12}, \cite{Chen:13}, \cite{Chu:13}, \cite{Zen:14}, \cite{Sangineto:14},
\cite{Kaltwang:12}:  at a time, data from 23 persons is used for training and from the 1 person for
testing.

Furthermore, a scenario where testing and training datasets are disjoint with respect to the person
is motivated by use--cases for emergency units and critically ill persons where it is not possible
to have neutral (i.e. without pain) images for the incoming patients. Thus, we consider that image
oriented k-fold scenarios, (e.g. in \cite{Rudovic:13}) are more theoretically oriented than
practically.

As the number of images with positive examples (with a specific AU or with Pain label) is much
lower than the one containing negative data, for the actual training the two sets were made even;
the chosen negative examples  were randomly selected. To increase the robustness of the system,
three classifiers were trained in parallel with independently drawn examples and the system output
was taken as the average of the classifiers.

For the actual discrimination of the pain intensity, we plied the same model as in the case of
similar works, \cite{Lucey:12}, \cite{Kaltwang:12}. We used two levels of classifiers (late fusion
scheme): first, each category of features was fed into the set of three Support Vector Regressors
(SVR) (with radial basis kernel function, cost 4 and $\Gamma =2^{-3.5}$). Landmarks were not
spectrally regressed (i.e were not re-represented with eq. (\ref{Eq:Proj}) ) but directly passed to
the SVRs. The results were fused together within a second level of boosted ensemble of four SVRs.
The implementation of the SVR is based on LibSVM  \cite{Chang:11}.

\subsection{Pain Estimation - Results}
The preferred implementation was by direct estimation of Prkachin - Solomon score of pain.
Alternatively, one may consider as intermediate step the AU estimation, followed by pain prediction
using equation (\ref{Eq:PrkSolomon}); yet previous research \cite{Kaltwang:12}, \cite{Florea:07}
showed that this method produces weaker results since errors are cumulated.

The best performing method for individual image based Prkachin-Solomon pain score estimation
produces a correlation coefficient of $\rho = 0.551$, and a mean square error of
$\overline{\varepsilon^2}=1.187$. The area under curve is $AUC = 80.9$. The best temporal filtering
increased the correlation to $\rho = 0.562$ and decreased the mean square error to
$\overline{\varepsilon^2}=0.885$. Best AUC achieved was $AUC=82.1$. The next subsection will
further detail these results and their implications.

Given a new UNBC image and the relevant landmarks positions, the query to determine the pain
intensity for that image takes approximately 0.15 seconds on a single thread Matlab implementation
on an Intel Xeon at 3.3 GHz. Temporal filtering adds a delay due to the consideration of a temporal
window around the current frame; this window is larger then a blink (which has a typical duration
of 300-400 milliseconds) and it adds a delay of $\approx$ 1 second.

\begin{figure*}[t]
    \begin{center}
        \begin{tabular}{cc}
            \includegraphics[width=0.48 \textwidth]{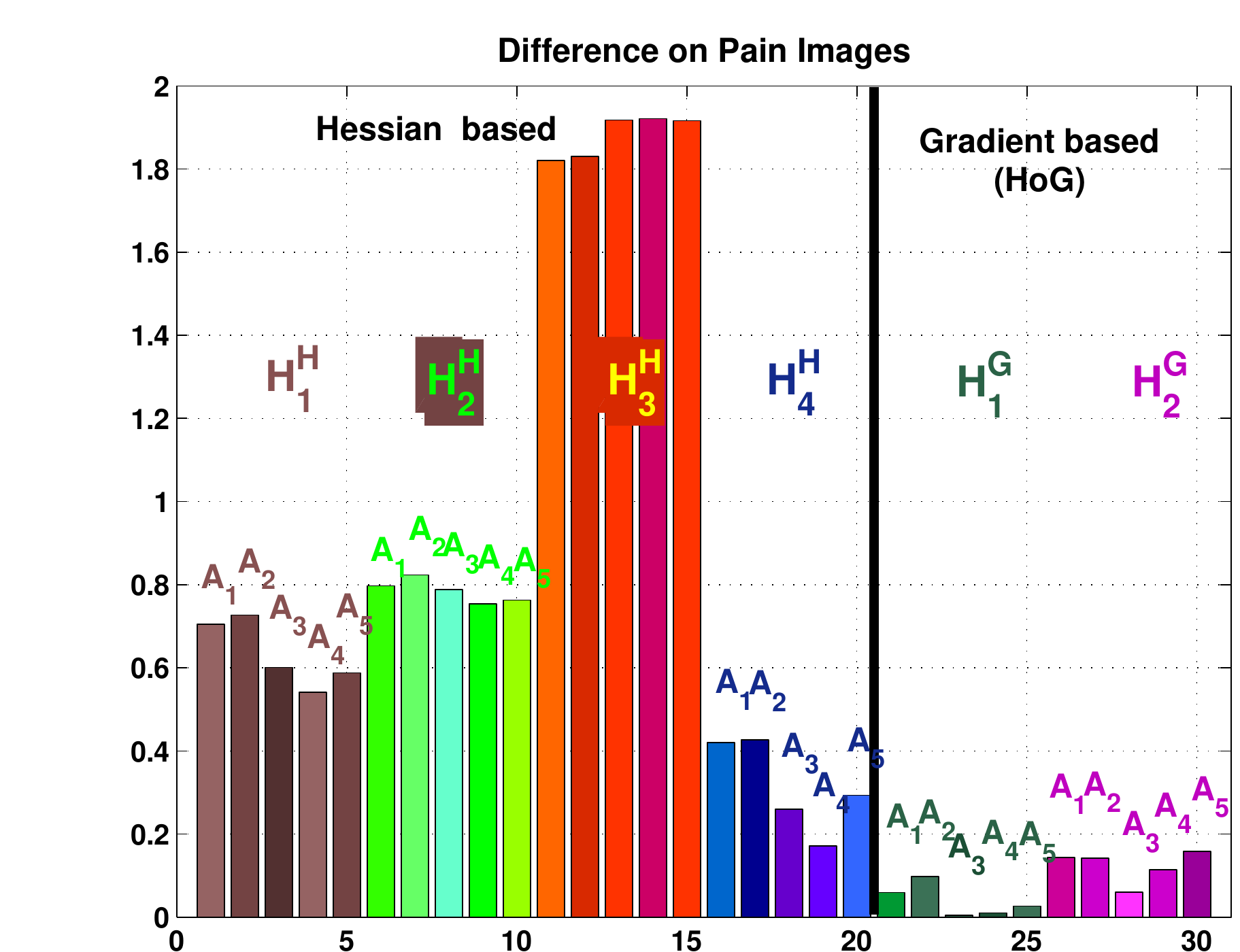} &
            \includegraphics[width=0.48 \textwidth]{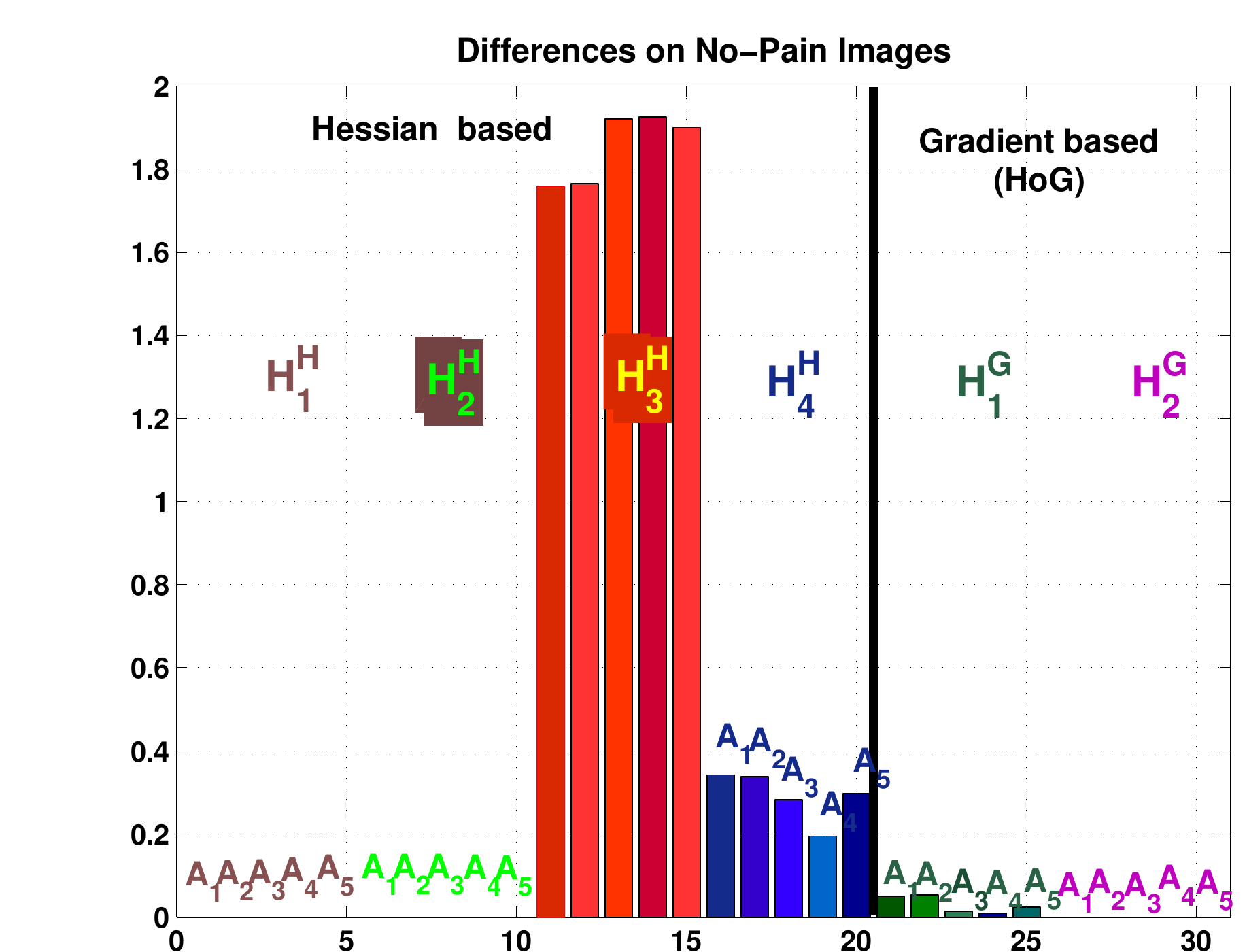}
        \end{tabular}
    \end{center}
    \caption{ Sum of absolute differences when comparing all images without pain and respectively with
    intense pain to a chosen no-pain reference image. Ideally, we aim for large values in the
    left plot and zeros in the right one. $A_1$ refers to the first area of interest (i.e. around the
    left eye), $A_2$ to the second one (around right eye), etc. \label{Fig:DiffsHoT}}
\end{figure*}

\subsection{Experiments}

\subsubsection{Are the HoT Features Really Useful?}
First, we investigate the capabilities of the \emph{HoT features} by considering the following
\emph{example}: we take the first frontal image without pain for each person and consider its HoT
features as reference; next, we compute the HoT features of all the images with a pain intensity
higher than 4 and of all the images without pain for each person separately. We plot the sum of
absolute differences between the set considered as reference to the mentioned images with and
without pain respectively. The results are presented in figure \ref{Fig:DiffsHoT}. Ideally, large
values are aimed in the left plot and zeros in the right one. We note that, for this particular
example, the largest contribution in discriminating between pain and no-pain cases was due to
Hessian based $H_1^H$ and $H_2^H$ histograms. Gradient based histograms lead to inconclusive
differences in the case of intense pain, while $H_3^H$ and $H_4^H$ produced large values also for
the no-pain case.

\begin{figure*}[t]
    \begin{center}
        \begin{tabular}{cc}
            \includegraphics[width=0.48 \textwidth]{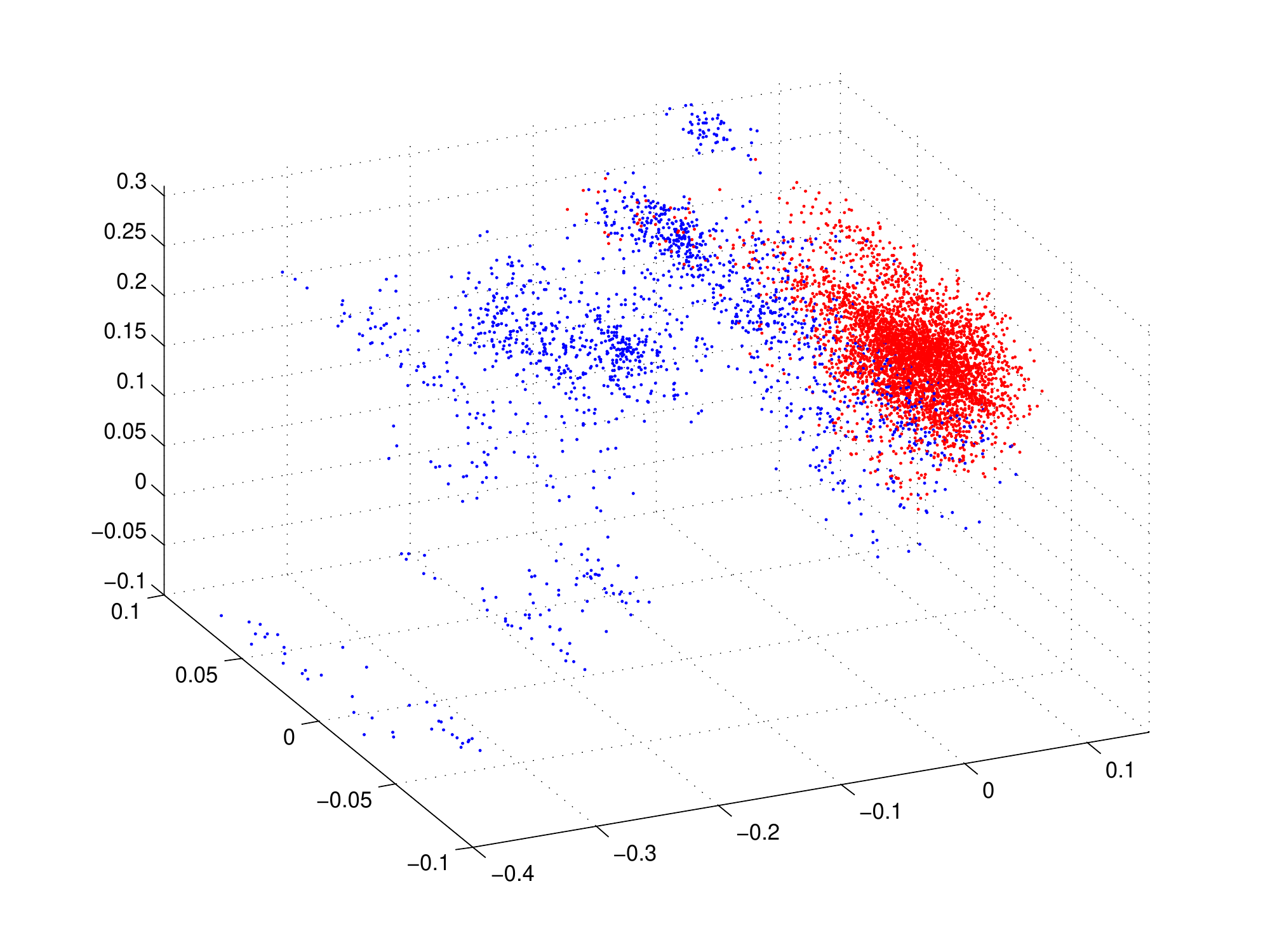} &
            \includegraphics[width=0.48 \textwidth]{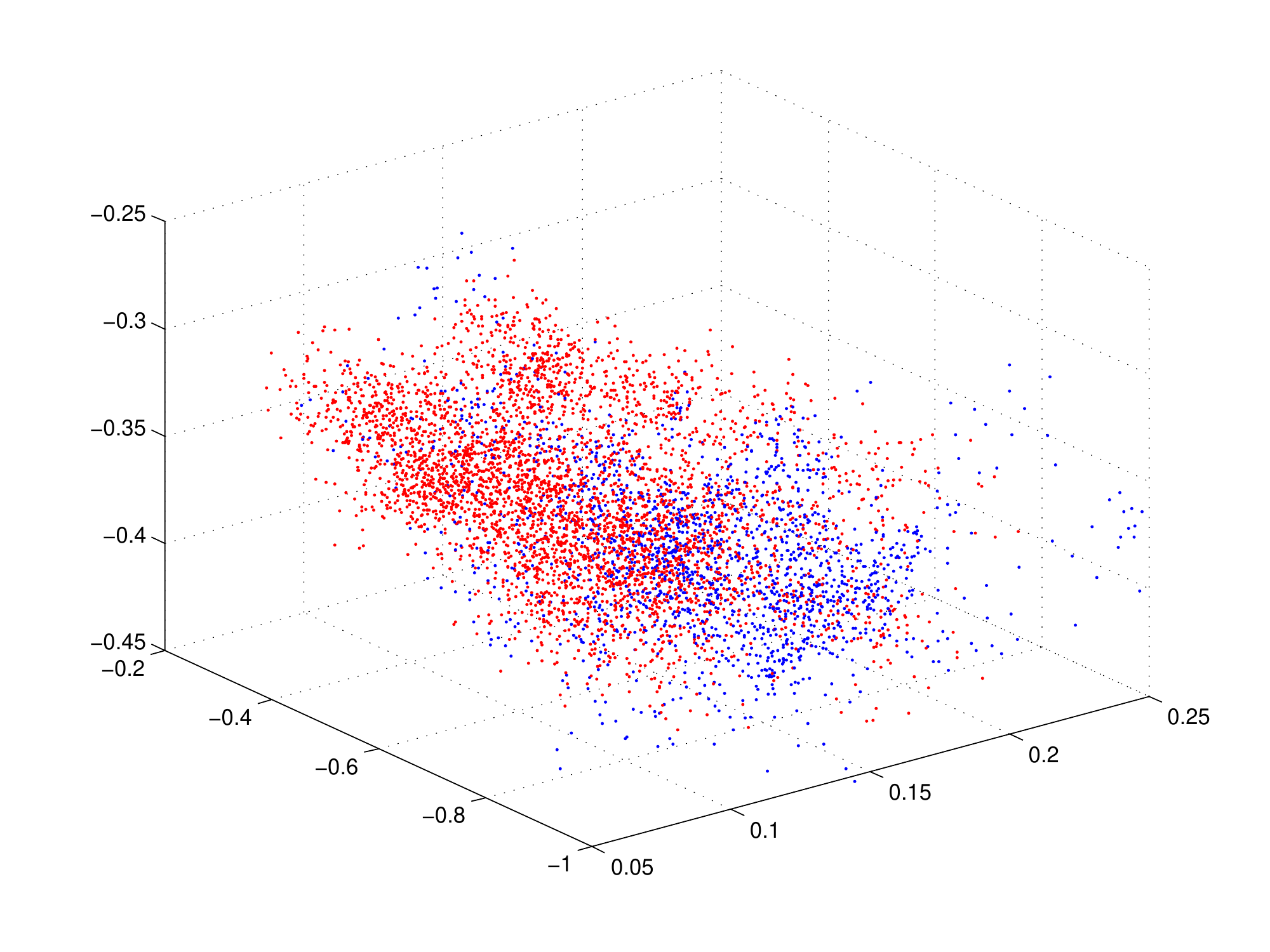}
        \end{tabular}
    \end{center}
    \caption{ Data clustering for Hessian based histograms (left) and respectively gradient based
        histograms (right). in each case the first three axes are retained. With red are frames with  high pain intensity and with blue are
        the first 4000 no-pain images.
        As one can see, the data is fairly clustered for Hessian based features and less for gradient based ones.
            \label{Fig:HoTClusterring}}
\end{figure*}

Furthermore, if considering the first 3 dimensions as selected by the transferred SR-M, the first
4000 no-pain and all the intense pain (i.e. higher than 4) cases are clustered as shown in figure
\ref{Fig:HoTClusterring}. The clusters in the Hessian based space are fairly visible suggesting
that: (1) HoT features are more powerful if they include Hessian based data, while addressing the
pain problem and (2) identification of high pain is doable. Yet, we did not plot the data
corresponding to low levels of pain which fills the intermediate space and, in fact, makes the
discrimination difficult.

\subsubsection{Feature Contribution}

To appraise  the overall contribution of each histogram type to the facial based pain intensity
estimation we present the results in table \ref{Tab:HoTResults}. To have a reference with respect
to state of the art features, we fill in with results achieved for the same problem by Kaltwang et
al. \cite{Kaltwang:12}. As one may see, if taken individually the proposed histograms under-perform
state of the art features. Yet different categories complement each other well and by combining
them  we obtain improved results.

\begin{table*}[t]
    \caption{Accuracy of pain intensity estimation using the Prkachin - Solomon formula. We report the
    achieved results for various versions of features used: containing only  Hessian based histograms
    ($H_i^H$ - Hess), only gradient based histograms ($H_i^G$ - Grad)  and  both of them  to form the
    so called Histogram of Topographical (HoT = Grad+Hess) features; the complete version contains
    landmarks (marked as PTS) and HoT. The relevant features were in each case learned with the
    modified version of Spectral Regression (SR-M) on the Cohn-Kanade database (CK) via self--taught learning.
    The Prkachin - Solomon score is estimated directly by the classifiers which were trained accordingly.
    \label{Tab:HoTResults}  }
    \begin{center}
    \begin{tabular}{|c||  c|c|c|c|| c|c|c|}
        \hline
        Work  & \multicolumn{4}{|c|}{ Proposed } &  \multicolumn{3}{|c|}{ \cite{Kaltwang:12} } \\
        \hline
        Feature &  Hess  &  Grad & HoT & HoT+PTS         &   PTS  & DCT  & LBP    \\ \hline
        Measure & \multicolumn{7}{|c|}{ Mean Square Error $\overline{\varepsilon^2}$} \\ \hline
                &  3.76  & 4.67  & 3.35 & 1.187       &  2.592   & 1.712    & 1.812      \\ \hline
        Measure & \multicolumn{7}{|c|}{Correlation, $\rho$} \\ \hline
                & 0.252  & 0.341 & 0.417  & 0.551        & 0.363    & 0.528    & 0.483      \\ \hline
    \end{tabular}
    \end{center}
\end{table*}

To detail the contribution of each histogram type, as defined in section \ref{Sect:Hot}, we remove
one type of histogram and see the overall effect over the pain score. In table
\ref{Tab:HistResults} we report the achieved relative accuracy obtained with only part of histogram
types. The decrease is larger for the more important types. Landmarks are skipped for this
experiment. As one can see, all the histograms contribute positively.

\begin{table*}[t]
    \caption{Contribution of each of the histogram types used. We report the Pearson correlation
    coefficient, $\rho$ when the mentioned type of histogram is removed. The reference is the right-most
    result (all histograms used). Thus, smaller is the value (i.e. larger is the decrease), higher is the contribution of the
    specific type of histogram. \label{Tab:HistResults}  }
    \begin{center}
        \begin{tabular}{|c|c|c|c|c|c|c|c|}
            \hline
            Histogram removed & $H_1^H$ & $H_2^H$ & $H_3^H$ & $H_4^H$ & $H_1^G$ & $H_2^G$ & None - HoT \\ \hline
            Correlation, $\rho$ & 0.331 & 0.368   &  0.355  & 0.358   &  0.351  & 0.192   & 0.417\\ \hline
        \end{tabular}
    \end{center}
\end{table*}

\subsubsection{Feature Selection and Transfer Learning }

In table \ref{Tab:FeatSelec_CK} we present the overall performance when various possibilities of
transfer learning are considered. The internal data representation may be perceived as unsupervised
feature selection. In this sense, beyond the proposed modified Spectral Regression (SR-M), we
tested the standard Spectral Regression (SR) \cite{Cai:07} and the  Locality Preserving Projection
(LPP) \cite{He:03} as it is the inspiration for SR. We also tested the standard Principal Component
Analysis as being the foremost dimensionality reduction method and its derivation through
Expectation--Maximization, namely Probabilistic PCA (PPCA), \cite{Tipping:99}; further, we included
the Factor Analysis (FA) as it is a generalization of PCA and a more recent derivation of the PCA:
the Robust PCA (RPCA) based on Mixture of Gaussian for the noise model (RPCA-MOG) \cite{Zhao:14},
which is an improvement over the standard RPCA introduced by Candes et al. \cite{Candes:11} that
uses Principal Component Pursuit to find a unique solution.

\begin{table*}[t]
    \caption{Accuracy of pain intensity estimation achieved results when self--taught learning
     (i.e. feature selection was learned on the Cohn-Kanade database and used on UNBC) with
     dimensionality reduction method. Details are in text accordingly.
     \label{Tab:FeatSelec_CK}  }
    \begin{center}
    \begin{tabular}{|c|| c||c|c| c|c|c|c|}
        \hline
        Feature &            SR-M  &  SR  & LPP  & PCA  & PPCA & RPCA-MOG & FA \\ \hline \hline
        Measure & \multicolumn{7}{|c|}{ Mean Square Error $\overline{\varepsilon^2}$} \\ \hline
                &  1.187  & 1.183 & 1.203 & 1.181 & 1.173 & 3.891 & 2.746      \\ \hline
        Measure & \multicolumn{7}{|c|}{Correlation, $\rho$} \\ \hline
                & 0.551   & 0.545 & 0.544 & 0.541 & 0.545 & 0.522 & 0.540   \\ \hline
    \end{tabular}
    \end{center}
\end{table*}

\begin{table}[t]
    \caption{Comparison of the achieved accuracy of pain intensity estimation when feature selection is
        learned on the Cohn-Kanade database (i.e. self--taught learning) or directly on the UNBC
                database (i.e. no transfer). \label{Tab:FeatSelect_UNBC}  }
    \begin{center}
    \begin{tabular}{|c|| c|c|| c|c|}
        \hline
        \begin{tabular}{c} Database \\ for learning\end{tabular}
         & \multicolumn{2}{|c|}{ Cohn - Kanade}  & \multicolumn{2}{|c|}{ UNBC    }   \\
         \hline \hline
        Feature &            SR-M  &  PPCA  & SR-M  & PPCA   \\ \hline \hline
        Measure & \multicolumn{4}{|c|}{ Mean Square Error $\overline{\varepsilon^2}$} \\ \hline
                &  1.187  & 1.173 & 1.203 & 1.181       \\ \hline
        Measure & \multicolumn{4}{|c|}{Correlation, $\rho$} \\ \hline
                & 0.551   & 0.545 & 0.532 & 0.532   \\ \hline
    \end{tabular}
    \end{center}
\end{table}

Other considered alternatives are to perform no transfer at all, or to extract the inner data
representation directly from the labelled UNBC database. The comparative results for these cases
are presented in table \ref{Tab:FeatSelect_UNBC}. The results show that specifically relying on the
adapted similarity measure (SR-M) and taking into account a larger number of persons, the
discrimination capability increases.

\begin{figure}[t]
    \begin{center}
            \includegraphics[width=0.4 \textwidth]{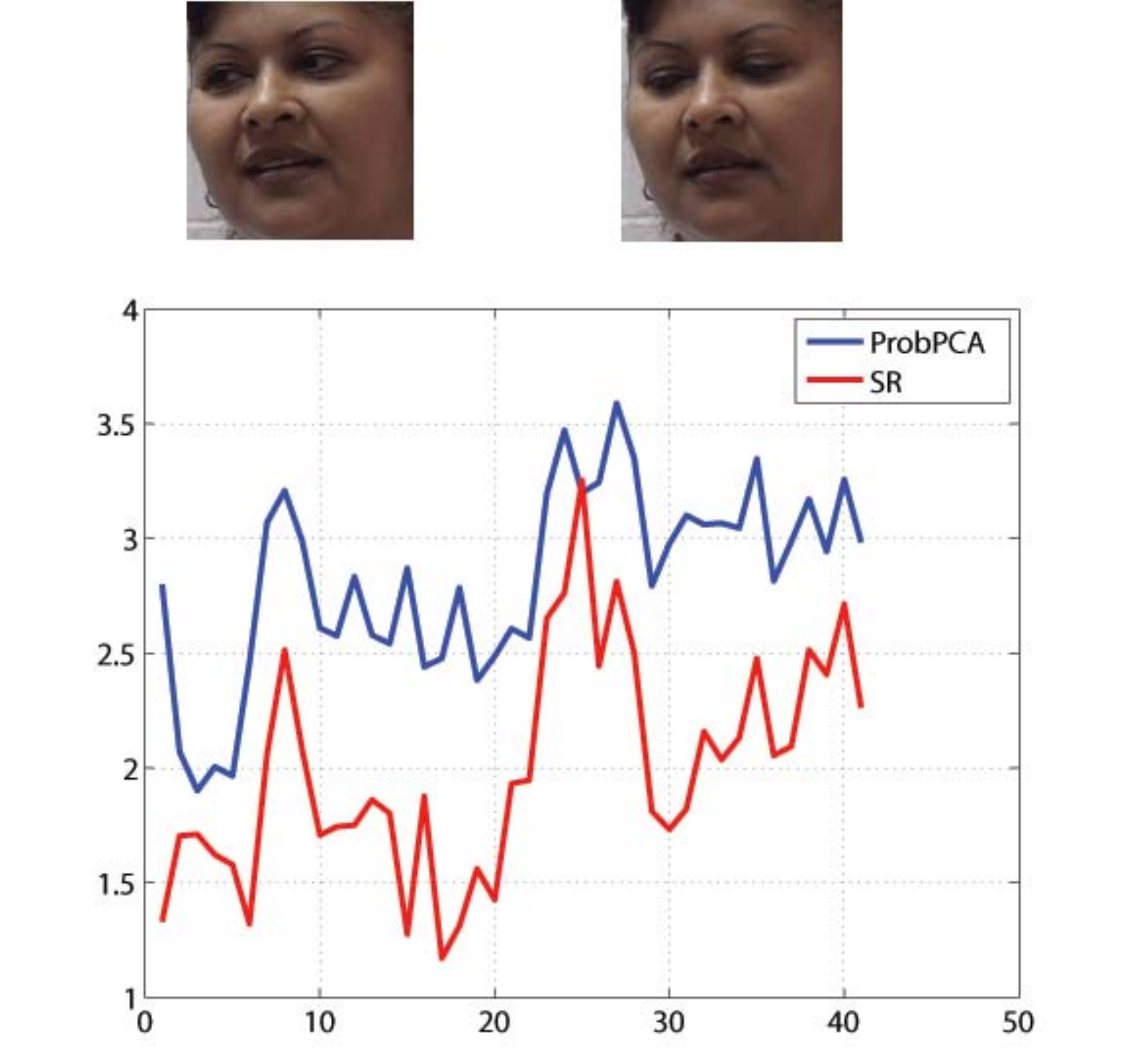}
    \end{center}
    \caption{A wave form for continuous pain intensity estimation taken from person 1 of the database.
    The red line is the pain estimation using Spectral Regression while the blue
    is with PPCA. The modes on SR plot are much more visible. \label{Fig:PCA_VS_SR}  }
\end{figure}

A numerical comparison between our modified version of spectral regression and the probabilistic
PCA, in transfer, shows little difference. Yet, we argue for the superiority of our method based on
analysis of the continuous pain intensity signals: the major difference is that our method shows a
bias towards blink and consistent results, the reduction based on PCA simply fails in some
situations without being able to make any correlation between them. A typical case is illustrated
in figure \ref{Fig:PCA_VS_SR}.

\subsubsection{Comparison with State of the Art for the Transfer Learning Procedure}
To give a quantitative comparison of the performance of the proposed self--taught learning method,
we note that multiple methods report transfer learning enhanced performances on the UNBC McMaster
Pain database. All of them applied the same evaluation procedure.

\citeyear{Chen:13}, \citeyear{Zen:14} and \citeyear{Sangineto:14} used histograms of LBP followed
by PCA reduction of dimensionality and various classification methods, by directly applying it to
training data or by relying on transductive transfer learning; \citeyear{Chen:13} report results
for AdaBoost and  Transductive Transfer AdaBoost (TTA); \citeyear{Chu:13} for Selective Transfer
Machine (STM);  \citeyear{Zen:14} report results for Transductive Support Vector Machine (TSVM) and
Support Vector-based Transductive Parameter Transfer (SVTPT) \cite{Zen:14}; Sangineto et al. for
Transductive Parameter Transfer with Density Estimate Kernel. As one may note, all the methods are
transductive transfer learning (i.e. the source and target tasks are the same, while the source and
target domains are different) while our method is part of the inductive transfer learning category
(i.e. the target task is different from the source task, no matter when the source and target
domains are the same or not \cite{Jialin:10}).

In table \ref{Tab:Comparison_AUC} we present the results reported by the mentioned works
comparatively to the performance of the proposed method. As one can see, our method reaches the
best accuracy.

\begin{table}[t]
    \centering
    \caption{Comparison with state of the art transfer learning methods using the achieved Area Under Curve (AUC).
    The explanation for the acronyms is in text.
        \label{Tab:Comparison_AUC} }
    \begin{tabular} { |c| c|c|c|c| c|c|c|c|}
        \hline
            Method & AUC \\ \hline
            AdaBoost \cite{Chen:13} & 76.9 \\ \hline
            TTA \cite{Chen:13}  & 76.5 \\ \hline
            TSVM  \cite{Zen:14} & 69.8 \\ \hline
            STM \\ \cite{Chu:13} & 76.8 \\ \hline
            TPT \\ \cite{Sangineto:14} & 76.7 \\ \hline
            SVTPT \\ \cite{Zen:14} & 78.4\\ \hline
            \textbf{Proposed}& \textbf{80.9} \\ \hline
    \end{tabular}
\end{table}


\subsection{Temporal Filtering}

The results achieved with the three methods of temporal filtering, given the still image pain
estimation are presented in table \ref{Tab:Temporal}.

\begin{table*}[t]
    \caption{Comparison of the achieved accuracy of pain intensity estimation when the three methods
         for temporal filtering were included: based on linear regression (LR), when the vicinity
         was a feature of MLP and with strict ordering description.
         \label{Tab:Temporal}  }
    \begin{center}
    \begin{tabular}{|c| c| c|c|c|c| }
        \hline
        Method &    Still  &  Temporal--LR  & Vicinity--MLP & Strict ordering--MLP  & Strict ordering--SVM        \\ \hline
        Measure & \multicolumn{5}{|c|}{ Mean Square Error $\overline{\varepsilon^2}$} \\ \hline
                &  1.187  & \textbf{0.885} & 1.137 & 1.200 & 1.280        \\ \hline
        Measure & \multicolumn{5}{|c|}{Correlation, $\rho$} \\ \hline
                & 0.551   & \textbf{0.562} & 0.535 & 0.529 & 0.558  \\ \hline
    \end{tabular}
    \end{center}
\end{table*}

While analyzing the results, all methods lead to improved mean square error and area under curve.
Regarding the correlation, from a quantitative point of view, the method based on linear regression
(LR), which has the main purpose of removing the noise in the estimated values, performs the best.
Yet this method is an incremental improvement of the still pain estimation.

The other filtering solutions produce, in fact, mixt results; while overall they indicate a
decrease or a small increase of the correlation, in fact they boost the performance of results on
half of the persons with more than 0.05, in average. The persons with increase are the ones where
the methods performed better than average (i.e. correlation was  $\rho>5$); here, the blinks were
correctly removed and the temporal filtered signal comes much closer to the ground truth. However,
on persons with below average initial results, the filtering de-correlates even more the estimated
values with respect to ground truth. These are persons that exhibit different pain faces, such as
opening the mouth (e.g. the person from the last column in figure \ref{Fig:UNBC_examples}) or
bowing the head to the low left.
Concluding, if either more data is available for learning or if the system is further robustified
with respect to the person, the machine learning temporal filtering will be more useful; now it is
a mere noise reduction does the work.


\subsection{Comparison with State of the Art.}

\begin{figure*}
\begin{center}
    \begin{tabular}{cc}
        \begin{tabular}{|c|c|c|}
            \hline
            Method & MSE, $\overline{\varepsilon^2}$ & Correlation, $\rho$ \\ \hline
            Proposed--Still    & 1.187 & 0.551 \\ \hline
            Proposed--Temporal & \textbf{0.885} & 0.562 \\ \hline
             \begin{tabular}{c} PTS+DCT \\ \cite{Kaltwang:12} \end{tabular} & 1.801 & 0.489 \\ \hline
             \begin{tabular}{c}PTS+LPB \\ \cite{Kaltwang:12}  \end{tabular} & 1.567 & 0.485 \\ \hline
             \begin{tabular}{c} PTS+DCT+LPB \\ \cite{Kaltwang:12} \end{tabular} & 1.386 & \textbf{0.590} \\ \hline
            \begin{tabular}{c} HoT+SR \\ \cite{Florea:14}  \end{tabular} & 1.183 & 0.545 \\ \hline
        \end{tabular} &
        \begin{tabular}{c}
              \includegraphics[width=0.4 \textwidth]{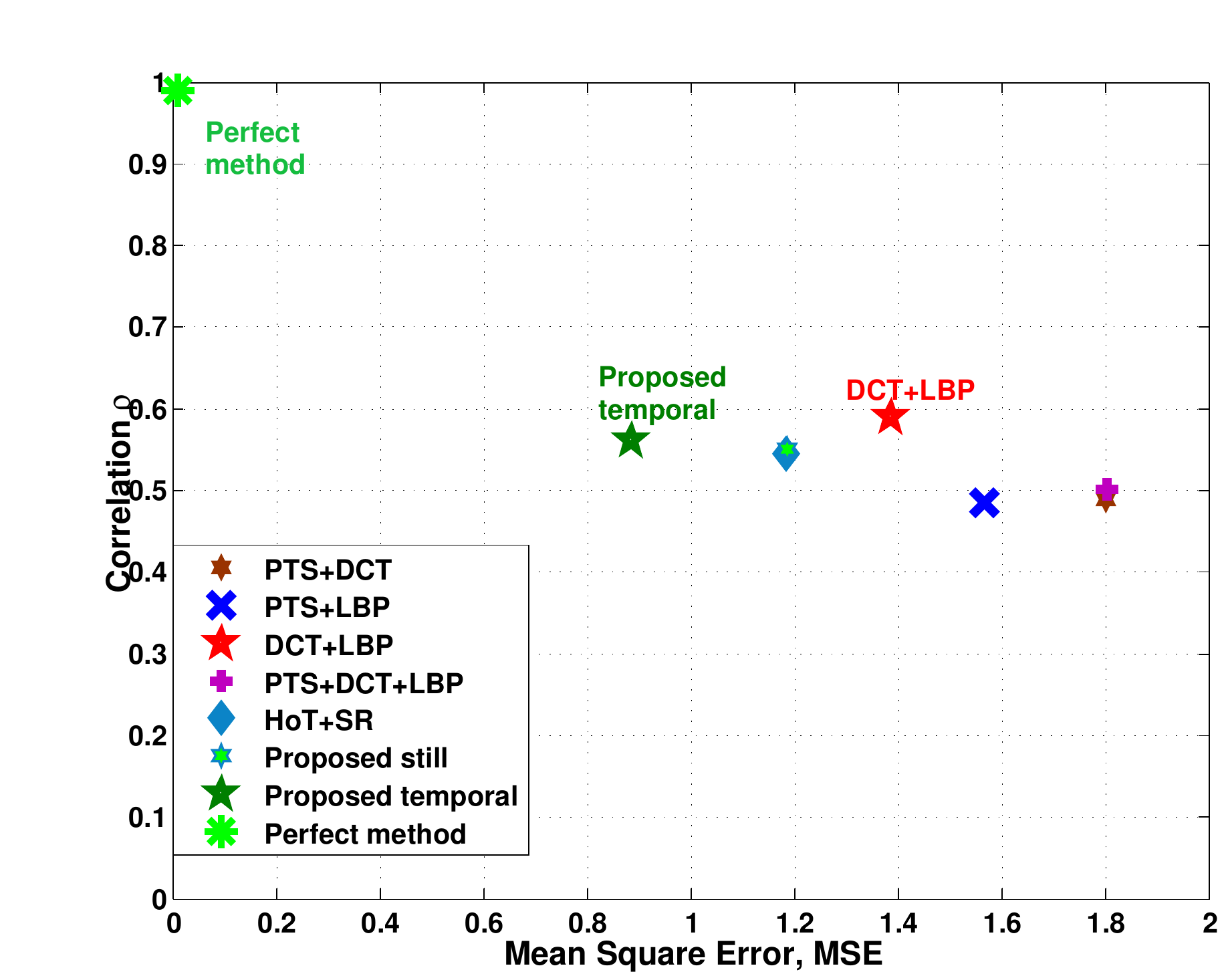}
        \end{tabular}
        \\
        (a) & (b)
    \end{tabular}
    \caption{(a) Numerical comparison of the achieved accuracy of pain intensity estimation with various state of the art methods.
     (b)Pearson correlation coefficient vs mean square error for the methods presented
    in the left-hand table. The perfect method is placed in the top left corner.
    \label{Fig:Comparison}}
\end{center}
\end{figure*}

As mentioned in section \ref{Subsect:priorArt}, there exist several methods reporting results on
the UNBC Pain database. Yet, only \citeyear{Kaltwang:12} and our previous work \cite{Florea:14}
tested on the entire database, with separation between persons when considering testing/training
folds and reported continuous pain intensity. The work from \cite{Kaltwang:12} consists in trying
several combination of feature coupled  with a Relevance Vector Machine (RVM- which is the SVM
reinterpreted under Bayesian framework) and fused with a second layer of RVM; we present all of
them to have a better comparison in the left hand table from figure \ref{Fig:Comparison}.


Mainly, the highest mean square error is obtained by the still image identification followed by
temporal filtering outperforming the next competitor by near 0.3 pain levels. In this category, it
is followed by our previous method \cite{Florea:14} and by the here proposed still image
estimation. Regarding the correlation coefficient, our methods set ranks second after the
combination of DCT with LPB fused by a RVM. Surprisingly, the direct combination of landmarks with
features reported by \cite{Kaltwang:12} does not lead to very good results.

Taking into account that there are different winners at different categories, to have a better
image of relation between them, we plotted the results from table (a) \ref{Fig:Comparison} as  MSE
vs $\rho$ axis (see figure \ref{Fig:Comparison} (b) ). In such a plot, a perfect method will have
$MSE=0$ and $\rho=1$ and it will be placed in the top left corner. As one can see, the proposed
temporal method is closer to the perfect one's position.


\section{Discussion and Conclusions}

In this paper we introduced the  Histogram of Topographic features to describe faces. The addition
of Hessian based terms allowed separation of various face movements and, thus, of pain intensity
levels. The robustness of the system was further enhanced by a new transfer learning method which
was inspired from the self--taught learning paradigm and relied on preserving the local similarity
of the feature vectors as learned over a more consistent database in terms of persons, to ensure
that relevant dimensions of the features are used in the subsequent classification process.

Regarding the addition of the actual features, while their individual contribution was rather small
when compared with consecrated features, they complemented each other well, as showed by the
increase of the overall performance when all feature types were used. As showed in table
\ref{Fig:Comparison} (a), this is not the case for features employed in previous solutions, which
argues for the consideration of the complete topographical description.

The transfer learning from a database with larger number of persons increased the system
robustness. More precisely, the solution that did not use the transfer procedure on some persons
lead to better results, with the cost of providing smaller accuracy on others that are more
different from the training faces. The transfer provided more consistent results overall, a fact
which was proved by the entropy of the correlation coefficient increase from 9.10 to 9.37,
enhancing the generalization with respect to person change. Such property is desirable taking into
account the different characteristics of pain expressivity, trait which impedes the temporal filter
to have an overall beneficial effect with results of noise reduction. Furthermore, the proposed
transfer learning method performs better when compared with similar attempts but based on
transductive transfer learning, as showed in table \ref{Tab:Comparison_AUC}.

The system provides indeed some failures. The AU 43 (closing eyes), according to eq.
(\ref{Eq:PrkSolomon}), contributes to pain intensity, not all blinks are pain-related; the system,
as in the case of \cite{Kaltwang:12}, mistakenly associate blinks of specific persons with pain.
Other failures are in cases where the person's method of expressing pain is rather different from
most of the others; for instance, the second person widely opens the eyes, instead of closing them,
leading the system to produce false negatives. Other errors are related to the fact that the person
is speaking during the test; false positives are associated with  persons bow (AU 54) or jerk (AU
58) the head while feeling pain; yet the behavior is not general. Still, while the effort of the
UNBC Pain Database creators was notable and made the foundation for advances on non-invasive pain
estimation from facial analysis, the database should be increased with more subjects to have
illustration of variability in pain faces.


\subsection{Continuation Paths}

At the end we consider that further research on the topic is beneficial and we would like to
emphasize several aspects, that in our opinion motivate such a necessity. First, the Prkachin -
Solomon score was found to be only moderately strong correlated with self--report (i.e. a Pearson
correlation coefficient of 0.66 or higher) \cite{Hammal:12} \cite{Prkachin:08}. Secondly, the
self-report was found to be the more accurate mean for appraisal of the pain intensity
\cite{Shavit:08}. Thirdly, the observational scores that were found to be more reliable, such as
the revised Adult Nonverbal Pain Scale (ANPS-R) and  the Critical Care Pain Observation Tool
(CPOT), contain additional indicators of pain such as the rigidness and the stiffness positions or
restless and excessive activity; these are gestures recognizable by a system for analysis of the
body posture. Concluding, additional data with annotation to inter-correlate the body  posture
estimation with facial pain assessment for facilitating further contribution on the topic of
automatic pain assessment, will make possible a gradual evolution to a fully developed, autonomous
system of assistive computer vision.


\section{Acknowledgements} The work has been partially funded by the Sectoral Operational Programme
Human Resources Development 2007-2013 of the Ministry of European Funds through the Financial
Agreement POSDRU/159/1.5/S/ 134398 and POSDRU/159/1.5/S/132395.

\bibliographystyle{icml2014}
\bibliography{HoT_Pain_Arxiv}

\end{document}